\documentclass{article}

\usepackage{times}
\usepackage{graphicx} 
\usepackage{subfigure}

\usepackage{natbib}

\usepackage{algorithm}
\usepackage{algorithmic}
\usepackage[algo2e,linesnumbered,lined,boxed,vlined]{algorithm2e}

\usepackage{amsfonts}
\usepackage{amssymb}

\newcommand{\indep}{\perp\!\!\!\perp}

\usepackage{hyperref}

\usepackage[accepted]{icml2016}

\usepackage{comment}
\usepackage{amsthm}
\usepackage{mathtools}
\usepackage{tabularx}
\usepackage{multirow}

\def\b{\ensuremath\boldsymbol}

\usepackage{lastpage}
\usepackage{fancyhdr}
\usepackage{url}
\icmltitlerunning{The Theory Behind Overfitting, Cross Validation, Regularization, Bagging, and Boosting: Tutorial}

\begin{document}

\twocolumn[
\icmltitle{The Theory Behind Overfitting, Cross Validation,\\ Regularization, Bagging, and Boosting: Tutorial}

\icmlauthor{Benyamin Ghojogh}{bghojogh@uwaterloo.ca}
\icmladdress{Department of Electrical and Computer Engineering, 
\\Machine Learning Laboratory, University of Waterloo, Waterloo, ON, Canada}
\icmlauthor{Mark Crowley}{mcrowley@uwaterloo.ca}
\icmladdress{Department of Electrical and Computer Engineering, 
\\Machine Learning Laboratory, University of Waterloo, Waterloo, ON, Canada
\\ \\
Lecture of Benyamin Ghojogh about this tutorial: \url{https://www.youtube.com/watch?v=wds4KdXQJIA} \\
Lecture of Prof. Ali Ghodsi about this tutorial: \url{https://www.youtube.com/watch?v=21jL0I6wbns}
}

\icmlkeywords{Tutorial, Locally Linear Embedding}

\vskip 0.3in
]

\begin{abstract}
In this tutorial paper, we first define mean squared error, variance, covariance, and bias of both random variables and classification/predictor models. Then, we formulate the true and generalization errors of the model for both training and validation/test instances where we make use of the Stein's Unbiased Risk Estimator (SURE). We define overfitting, underfitting, and generalization using the obtained true and generalization errors. We introduce cross validation and two well-known examples which are $K$-fold and leave-one-out cross validations. We briefly introduce generalized cross validation and then move on to regularization where we use the SURE again. We work on both $\ell_2$ and $\ell_1$ norm regularizations. Then, we show that bootstrap aggregating (bagging) reduces the variance of estimation. Boosting, specifically AdaBoost, is introduced and it is explained as both an additive model and a maximum margin model, i.e., Support Vector Machine (SVM). The upper bound on the generalization error of boosting is also provided to show why boosting prevents from overfitting. As examples of regularization, the theory of ridge and lasso regressions, weight decay, noise injection to input/weights, and early stopping are explained. Random forest, dropout, histogram of oriented gradients, and single shot multi-box detector are explained as examples of bagging in machine learning and computer vision. Finally, boosting tree and SVM models are mentioned as examples of boosting.
\end{abstract}

\begin{figure*}[!t]
\centering
\includegraphics[width=5in]{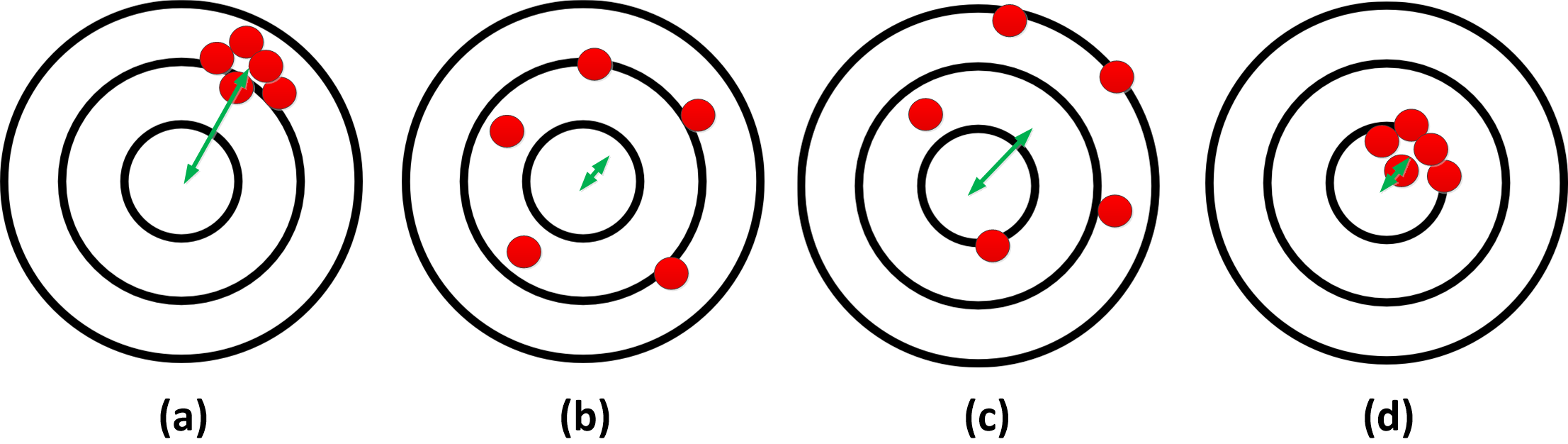}
\caption{The dart example for (a) high bias and low variance, (b) low bias and high variance, (c) high bias and high variance, and (d) low bias and low variance. The worst and best cases are (c) and (d), respectively. The center of the circles is the true value of the variable.}
\label{figure_dart}
\end{figure*}

\section{Introduction}

Assume we have a dataset of \textit{instances} $\{(\b{x}_i, y_i)\}_{i=1}^N$ with sample size $N$ and dimensionality $\b{x}_i  \in \mathbb{R}^d$ and $y_i \in \mathbb{R}$. 
The $\{\b{x}_i\}_{i=1}^N$ are the input data to the model and the $\{y_i\}_{i=1}^N$ are the observations (labels).
We denote the dataset by $\mathcal{D}$ so that $N := |\mathcal{D}|$. This dataset is the union of the disjoint subsets, i.e., training set $\mathcal{T}$ and test set $\mathcal{R}$; therefore:
\begin{align}
&\mathcal{D} = \mathcal{T} \cup \mathcal{R}, \\
&\mathcal{T} \cap \mathcal{R} = \varnothing.
\end{align}
For the training set, the observations (labels), $y_i$'s, are available. Although for the test set, we might also have $y_i$'s, but we do not use them for training the model. The observations are continuous or come from a finite discrete set of values in classification and prediction (regression) tasks, respectively.
Assume the sample size of training and test sets are $n := |\mathcal{T}|$ and $m := N-n$, respectively; therefore, we have $\{(\b{x}_i, y_i)\}_{i=1}^n$ as the training set.
In some cases where we want to have validation set $\mathcal{V}$ as well, the datasets includes three disjoint subsets:
\begin{align}
&\mathcal{D} = \mathcal{T} \cup \mathcal{R} \cup \mathcal{V}, \\
&\mathcal{T} \cap \mathcal{R} = \varnothing, \mathcal{T} \cap \mathcal{V} = \varnothing, \mathcal{V} \cap \mathcal{R} = \varnothing. 
\end{align}
We will define the intuitions of training, test, and validation sets later in this paper.

In this paper, we introduce overfitting, cross validation, generalized cross validation, regularization, bagging, and boosting and explain why they work theoretically. 
We also provide some examples of these methods in machine learning and computer vision.

It is noteworthy that Sections \ref{section_mse}, \ref{section_overfitting}, \ref{section_CV_theory}, and some parts of Section \ref{section_regularization} (i.e., analysis of overfitting and regularization using SURE), and Section \ref{section_bagging_theory} (analysis of bagging), were primarily proposed by Prof. Ali Ghodsi verbally in his lectures, at University of Waterloo, available on YouTube\footnote{See \url{https://www.youtube.com/watch?v=21jL0I6wbns}.}. The credit of those sections is his. 
Moreover some parts of Section \ref{section_regularization} have been discussed in the books \cite{friedman2001elements} and \cite{goodfellow2016deep}.

\section{Mean Squared Error, Variance, and Bias}

\subsection{Measures for a Random Variable}

Assume we have variable $X$ and we estimate it. Let the random variable $\widehat{X}$ denote the estimate of $X$. The \textit{variance} of estimating this random variable is defined as:
\begin{align}\label{equation_variance}
\mathbb{V}\text{ar}(\widehat{X}) := \mathbb{E}\big((\widehat{X} - \mathbb{E}(\widehat{X}))^2\big),
\end{align}
which means average deviation of $\widehat{X}$ from the mean of our estimate, $\mathbb{E}(\widehat{X})$, where the deviation is squared for symmetry of difference.
This variance can be restated as:
\begin{align}
\mathbb{V}\text{ar}(\widehat{X}) &= \mathbb{E}\big( \widehat{X}^2 + (\mathbb{E}(\widehat{X}))^2 - 2\widehat{X}\mathbb{E}(\widehat{X}) \big) \nonumber \\
&\overset{(a)}{=} \mathbb{E}(\widehat{X}^2) + (\mathbb{E}(\widehat{X}))^2 - 2\mathbb{E}(\widehat{X})\mathbb{E}(\widehat{X}) \nonumber \\
&= \mathbb{E}(\widehat{X}^2) - (\mathbb{E}(\widehat{X}))^2, \label{equation_variance_2}
\end{align}
where $(a)$ is because expectation is a linear operator and $\mathbb{E}(\widehat{X})$ is not a random variable.

Our estimation can have a bias. The \textit{bias} of our estimate is defined as:
\begin{align}\label{equation_bias}
\mathbb{B}\text{ias}(\widehat{X}) := \mathbb{E}(\widehat{X}) - X,
\end{align}
which means how much the mean of our estimate deviates from the original $X$.

The \textit{Mean Squared Error (MSE)} of our estimate, $\widehat{X}$, is defined as:
\begin{align}\label{equation_MSE}
\text{MSE}(\widehat{X}) := \mathbb{E}\big((\widehat{X} - X)^2\big),
\end{align}
which means how much our estimate deviates from the original $X$.

The intuition of bias, variance, and MSE is illustrated in Fig. \ref{figure_dart} where the estimations are like a dart game. We have four cases with low/high values of bias and variance which are depicted in this figure.

The relation of MSE, variance, and bias is as follows:
\begin{align}
&\text{MSE}(\widehat{X}) = \mathbb{E}\big((\widehat{X} - X)^2\big) \nonumber \\
&= \mathbb{E}\big((\widehat{X} - \mathbb{E}(\widehat{X}) + \mathbb{E}(\widehat{X}) - X)^2\big) \nonumber \\
&= \mathbb{E}\big((\widehat{X} - \mathbb{E}(\widehat{X}))^2 + (\mathbb{E}(\widehat{X}) - X)^2 \nonumber \\
&~~~~ + 2 (\widehat{X} - \mathbb{E}(\widehat{X})) (\mathbb{E}(\widehat{X}) - X) \big) \nonumber \\
&\overset{(a)}{=} \mathbb{E}\big( (\widehat{X} - \mathbb{E}(\widehat{X}))^2 \big) + (\mathbb{E}(\widehat{X}) - X)^2 \nonumber \\
&~~~~ + 2 \underbrace{(\mathbb{E}(\widehat{X}) - \mathbb{E}(\widehat{X}))}_{0} (\mathbb{E}(\widehat{X}) - X) \nonumber \\
&\overset{(b)}{=} \mathbb{V}\text{ar}(\widehat{X}) + (\mathbb{B}\text{ias}(\widehat{X}))^2, \label{equation_relation_MSE_variance_bias}
\end{align}
where $(a)$ is because expectation is a linear operator and $X$ and $\mathbb{E}(\widehat{X})$ are not random, and $(b)$ is because of Eqs. (\ref{equation_variance}) and (\ref{equation_bias}).

If we have two random variables $\widehat{X}$ and $\widehat{Y}$, we can say:
\begin{align}
&\mathbb{V}\text{ar}(a\widehat{X} + b\widehat{Y}) \overset{(\ref{equation_variance_2})}{=} \mathbb{E}\big((a\widehat{X} + b\widehat{Y})^2\big) - \big(\mathbb{E}(a\widehat{X} + b\widehat{Y})\big)^2 \nonumber \\
&\overset{(a)}{=} a^2\, \mathbb{E}(\widehat{X}^2) + b^2\, \mathbb{E}(\widehat{Y}^2) + 2ab\, \mathbb{E}(\widehat{X}\widehat{Y}) \nonumber \\
&~~~~ - a^2\, (\mathbb{E}(\widehat{X}))^2 - b^2\, (\mathbb{E}(\widehat{Y}))^2 - 2ab\, \mathbb{E}(\widehat{Y})\mathbb{E}(\widehat{Y}) \nonumber \\
&\overset{(\ref{equation_variance_2})}{=} a^2\, \mathbb{V}\text{ar}(\widehat{X}) + b^2\, \mathbb{V}\text{ar}(\widehat{X}) + 2ab\, \mathbb{C}\text{ov}(\widehat{X},\widehat{Y}), \label{equation_variance_of_two_variables}
\end{align}
where $(a)$ is because of linearity of expectation and the $\mathbb{C}\text{ov}(\widehat{X},\widehat{Y})$ is \textit{covariance} defined as:
\begin{align}\label{equation_covariance}
\mathbb{C}\text{ov}(\widehat{X},\widehat{Y}) := \mathbb{E}(\widehat{X}\widehat{Y}) - \mathbb{E}(\widehat{X})\,\mathbb{E}(\widehat{Y}).
\end{align}
If the two random variables are independent, i.e., $X \indep Y$, we have:
\begin{align}
&\mathbb{E}(\widehat{X}\widehat{Y}) \overset{(a)}{=} \int\!\!\! \int \widehat{x} \widehat{y} f(\widehat{x}, \widehat{y}) d\widehat{x} d\widehat{y} \overset{\indep}{=} \int\!\!\! \int \widehat{x} \widehat{y} f(\widehat{x}) f(\widehat{y}) d\widehat{x} d\widehat{y} \nonumber \\
&= \int \widehat{y} f(\widehat{y}) \underbrace{\int \widehat{x} f(\widehat{x}) d\widehat{x}}_{\mathbb{E}(\widehat{X})} d\widehat{y} = \mathbb{E}(\widehat{X}) \underbrace{\int \widehat{y} f(\widehat{y}) d\widehat{y}}_{\mathbb{E}(\widehat{Y})}  \nonumber \\
&= \mathbb{E}(\widehat{X})\, \mathbb{E}(\widehat{Y}) \implies \mathbb{C}\text{ov}(\widehat{X},\widehat{Y}) = 0, \label{equation_expectation_independent}
\end{align}
where $(a)$ is according to definition of expectation. 
Note that Eq. (\ref{equation_expectation_independent}) is not true for the reverse implication (we can prove by counterexample). 

We can extend Eqs. (\ref{equation_variance_of_two_variables}) and (\ref{equation_covariance}) to multiple random variables:
\begin{align}
&\mathbb{V}\text{ar}\Big(\sum_{i=1}^k a_i X_i \Big) \nonumber \\
&~~~~~~~ = \sum_{i=1}^k a_i^2\, \mathbb{V}\text{ar}(X_i) + \sum_{i=1}^k \sum_{j=1, j\neq i}^k a_i a_j \mathbb{C}\text{ov}(X_i, X_j), \label{equation_variance_multiple} \\
&\mathbb{C}\text{ov}\Big( \sum_{i=1}^{k_1} a_i X_i, \sum_{j=1}^{k_2} b_j Y_j \Big) = \sum_{i=1}^{k_1} \sum_{j=1}^{k_2} a_i\, b_j\, \mathbb{C}\text{ov}(X_i, Y_j),
\end{align}
where $a_i$'s and $b_j$'s are not random.

\subsection{Measures for a Model}

\begin{figure}[!t]
\centering
\includegraphics[width=3in]{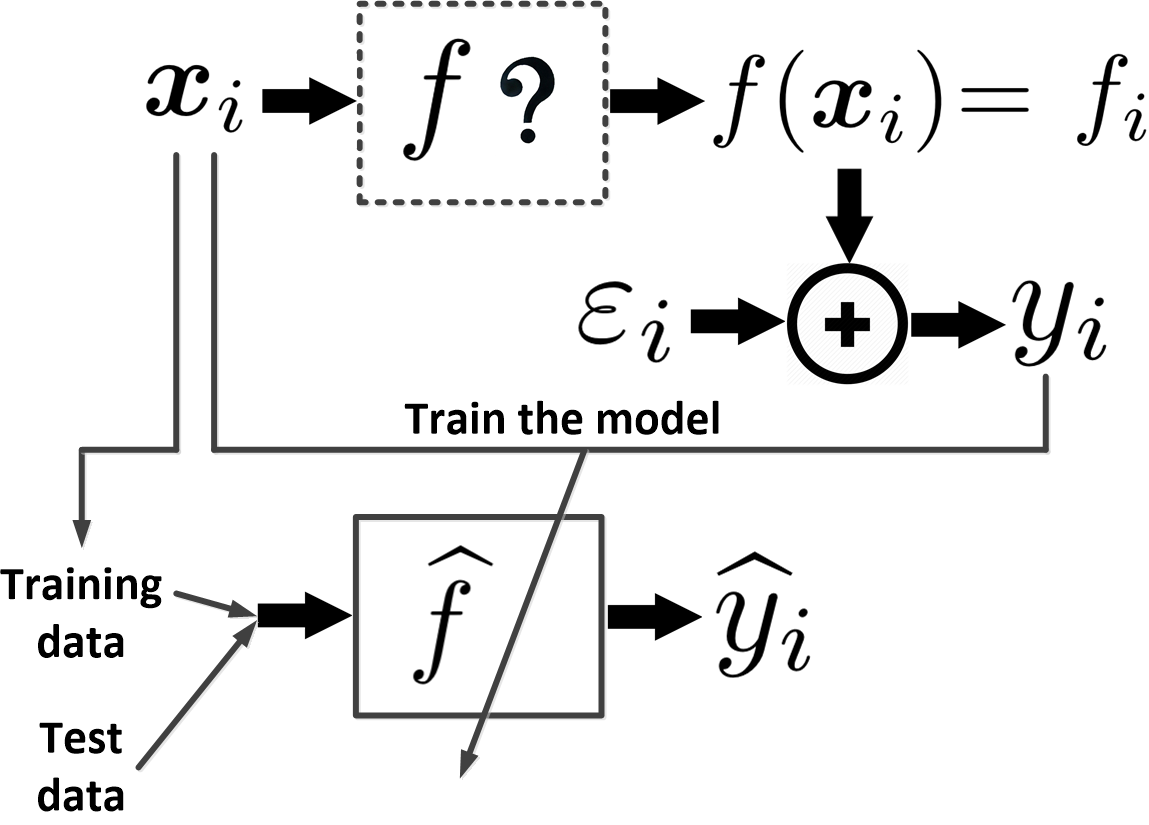}
\caption{The true model and the estimated model which is trained using the input training data and their observations. The observations are obtained from the outputs of the true model fed with the training input but corrupted by the noise. After the model is trained, it can be used to estimate the observation for either training or test input data.}
\label{figure_model}
\end{figure}

Assume we have a \textit{function} $f$ which gets the $i$-th input $\b{x}_i$ and outputs $f_i = f(\b{x}_i)$. Figure \ref{figure_model} shows this function and its input and output.
We wish to know the function which we call it the \textit{true model} but we do not have access to it as it is unknown. Also, the pure outputs (true observations), $f_i$'s, are not available. The output may be corrupted with an additive noise $\varepsilon_i$:
\begin{align}\label{equation_y_f_varepsilon}
y_i = f_i + \varepsilon_i,
\end{align}
where the noise is $\varepsilon_i \sim \mathcal{N}(0, \sigma^2)$. Therefore:
\begin{align}\label{equation_noise_mean_variance}
\mathbb{E}(\varepsilon_i) = 0, ~~~~ \mathbb{E}(\varepsilon_i^2) \overset{(\ref{equation_variance_2})}{=} \mathbb{V}\text{ar}(\varepsilon_i) + (\mathbb{E}(\varepsilon_i))^2 = \sigma^2, 
\end{align}
The true observation $f_i$ is not random, thus:
\begin{align}\label{equation_expectationOf_model}
\mathbb{E}(f_i) = f_i.
\end{align}
The input training data $\{\b{x}_i\}_{i=1}^n$ and their corrupted observations $\{y_i\}_{i=1}^n$ are available to us. We would like to approximate (estimate) the true model by a \textit{model} $\widehat{f}$ in order to estimate the observations $\{y_i\}_{i=1}^n$ from the input $\{\b{x}_i\}_{i=1}^n$. Calling the estimated observations by $\{\widehat{y}_i\}_{i=1}^n$, we want the $\{\widehat{y}_i\}_{i=1}^n$ to be as close as possible to $\{y_i\}_{i=1}^n$ for the training input data $\{\b{x}_i\}_{i=1}^n$. We train the model using the training data in order to estimate the true model. After training the model, it can be used to estimate the output of the model for both the training input $\{\b{x}_i\}_{i=1}^n$ and the unseen test input $\{\b{x}_i\}_{i=1}^m$ to have the estimates $\{\widehat{y}_i\}_{i=1}^n$ and $\{\widehat{y}_i\}_{i=1}^m$, respectively. The explained details are illustrated in Fig. \ref{figure_model}.

In this work, we denote the estimation of the observation of the $i$-th instance with either $\widehat{y}_i$ or $\widehat{f}_i$. 
The model can be a \textit{regression (prediction)} or \textit{classification} model. In regression, the model's estimation is continuous while in classification, the estimation is a member of a discrete set of possible observations.

\begin{figure}[!t]
\centering
\includegraphics[width=1.7in]{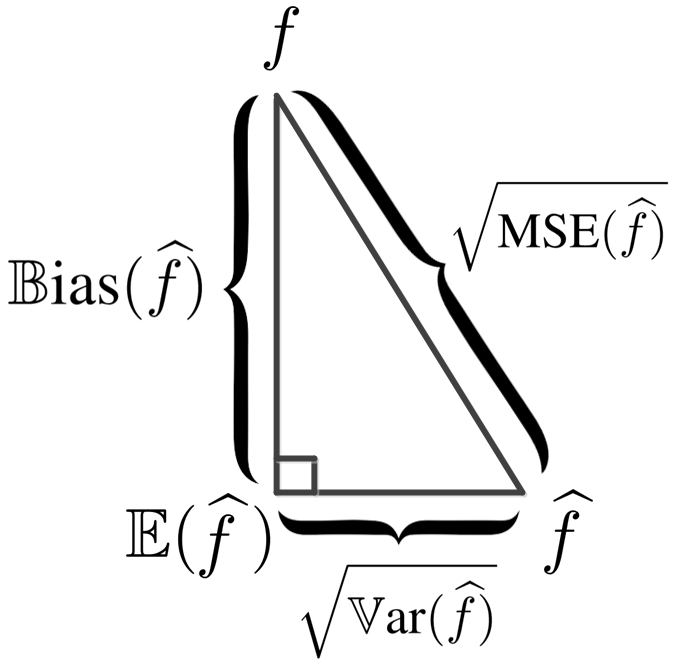}
\caption{The triangle of variance, bias, and MSE. The $f$ and $\widehat{f}$ are the true and estimated models, respectively.}
\label{figure_triangle_model}
\end{figure}

The definitions of variance, bias, and MSE, i.e., Eqs. (\ref{equation_variance}), (\ref{equation_bias}), and (\ref{equation_MSE}), can also be used for the estimation $\widehat{f}_i$ of the true model $f_i$. 
The Eq. (\ref{equation_relation_MSE_variance_bias}) can be illustrated for the model $f$ as in Fig. \ref{figure_triangle_model} which holds because of Pythagorean theorem. 

\subsection{Measures for Ensemble of Models}

If we have an ensemble of models \cite{polikar2012ensemble}, we can have some similar definitions of bias and variance (e.g., see the Appendix C in \cite{schapire1998boosting}).
Here, we assume the models are classifiers. 

If $\mathbb{P}(.)$ denotes the probability, the \textit{expected error} or \textit{prediction error} (PE) of the model $f_i$ is defined as:
\begin{align}
\text{PE}(f) := \mathbb{P}(\widehat{f}_i \neq y_i),
\end{align}
where $\widehat{f}_i$ is the estimation of trained model for the observation $y_i$ (input $\b{x}_i$).

In the parentheses, it is required to mention that Bayesian classifier is the optimal classifier because it can be seen as an ensemble of hypotheses (models) in the hypothesis (model) space and no other ensemble of hypotheses can outperform it (see Chapter 6, Page 175 in \cite{mitchell1997machine}). In the literature, it is referred to as \textit{Bayes optimal classifier}. However, implementing Bayesian classifier is difficult so they approximate it by \textit{naive Bayes} \cite{zhang2004optimality}. 

Back to our main discussion, let $\widehat{f}^*$ denote the Bayes optimal prediction. 
Also, let the estimate of each of the trained models by an ensemble learning method (such as bagging or boosting which will be introduced later) be denoted by $\widehat{f}$ which is trained using $\{(\b{x}_i, y_i)\}_{i=1}^n$. Finally, let $\widehat{f}^m$ denote the classification using majority voting between the models.
The bias and variance of the model can be defined as \cite{kong1995error}:
\begin{align}
&\mathbb{B}\text{ias}(\widehat{f}) := \text{PE}(\widehat{f}^m) - \text{PE}(\widehat{f}^*), \\
&\mathbb{V}\text{ar}(\widehat{f}) := \mathbb{E}(\text{PE}(\widehat{f})) -  \text{PE}(\widehat{f}^m).
\end{align}

There also exist another definition in the literature.
Suppose the sample space of data is the union of two disjoint subsets $\mathcal{U}$ and $\mathcal{B}$ which are the unbiased and biased sets with $\widehat{f}_i^m = \widehat{f}_i^*$ and $\widehat{f}_i^m \neq \widehat{f}_i^*$, respectively. We can define \cite{breiman1998arcing}:
\begin{align}
&\mathbb{B}\text{ias}(\widehat{f}_i) \nonumber \\
&:= \mathbb{P}(\widehat{f}_i^* = y_i, \b{x}_i \in \mathcal{B}) - \mathbb{E}(\mathbb{P}(\widehat{f}_i = y_i, \b{x}_i \in \mathcal{B})), \\
&\mathbb{V}\text{ar}(\widehat{f}_i) \nonumber \\
&:= \mathbb{P}(\widehat{f}_i^* = y_i, \b{x}_i \in \mathcal{U}) - \mathbb{E}(\mathbb{P}(\widehat{f}_i = y_i, \b{x}_i \in \mathcal{U})).
\end{align}

\section{Mean Squared Error of the Estimation of Observations}\label{section_mse}

Suppose we have an instance $(\b{x}_0, y_0)$. This instance can be either a training or test/validation instance. We will cover both cases.
According to Eq. (\ref{equation_y_f_varepsilon}), the observation $y_0$ is:
\begin{align}\label{equation_y0}
y_0 = f_0 + \varepsilon_0.
\end{align}
Assume the model's estimation of $y_0$ is $\widehat{f}_0$.
According to Eq. (\ref{equation_MSE}), the MSE of the estimation is:
\begin{align}
\mathbb{E}\big(&(\widehat{f}_0 - y_0)^2 \big) \overset{(\ref{equation_y0})}{=} \mathbb{E}\big((\widehat{f}_0 - f_0 - \varepsilon_0)^2 \big) \nonumber \\
&= \mathbb{E}\big((\widehat{f}_0 - f_0)^2 + \varepsilon_0^2 -2\,\varepsilon_0 (\widehat{f}_0 - f_0) \big) \nonumber \\
&= \mathbb{E}\big((\widehat{f}_0 - f_0)^2\big) + \mathbb{E}(\varepsilon_0^2) -2\,\mathbb{E}\big(\varepsilon_0 (\widehat{f}_0 - f_0) \big) \nonumber \\ 
&\overset{(\ref{equation_noise_mean_variance})}{=} \mathbb{E}\big((\widehat{f}_0 - f_0)^2\big) + \sigma^2 -2\,\mathbb{E}\big(\varepsilon_0 (\widehat{f}_0 - f_0) \big). \label{equation_MSE_y0}
\end{align}
The last term is:
\begin{align}
\mathbb{E}\big(\varepsilon_0 (\widehat{f}_0 - f_0) \big) \overset{(\ref{equation_y0})}{=} \mathbb{E}\big( (y_0 - f_0) (\widehat{f}_0 - f_0) \big).
\end{align}
For calculation of this last term, we consider two cases: (I) whether the instance $(\b{x}_0, y_0)$ is in the training set or (II) not in the training set. In other words, whether the instance was used to train the model (estimator) or not.

\subsection{Case I: Instance not in the Training Set}

Assume the instance $(\b{x}_0, y_0)$ was not in the training set, i.e., it was not used for training the model.
In other words, we have $y_0 \notin \mathcal{T}$.
This means that the estimation $\widehat{f}_0$ is independent of the observation $y_0$ because the observation was not used to train the model but the estimation is obtained from the model.
Therefore:
\begin{align*}
&\therefore ~~~ y_0 \indep \widehat{f}_0 \implies (y_0 - f_0) \indep (\widehat{f}_0 - f_0) \\
&\implies \mathbb{E}\big( (y_0 - f_0) (\widehat{f}_0 - f_0) \big) \\
&\overset{(a)}{=} \mathbb{E}\big( (y_0 - f_0) \big) \, \mathbb{E}\big( (\widehat{f}_0 - f_0) \big) \overset{(b)}{=} 0 \times \mathbb{E}\big( (\widehat{f}_0 - f_0) \big) = 0,
\end{align*}
where $(a)$ is because $(y_0 - f_0) \indep (\widehat{f}_0 - f_0)$ and $(b)$ is because:
\begin{align*}
\mathbb{E}\big( (y_0 - f_0) \big) = \mathbb{E}(y_0) - \mathbb{E}(f_0) \overset{(c)}{=} f_0 - f_0 = 0,
\end{align*}
where $(c)$ is because of Eq. (\ref{equation_expectationOf_model}) and:
\begin{align*}
\mathbb{E}(y_0) \overset{(\ref{equation_y0})}{=} \mathbb{E}(f_0) + \mathbb{E}(\varepsilon_0) = f_0 + 0 = f_0.
\end{align*}
Therefore, in this case, the last term in Eq. (\ref{equation_MSE_y0}) is zero. Thus:
\begin{align}
\mathbb{E}\big(&(\widehat{f}_0 - y_0)^2 \big) = \mathbb{E}\big((\widehat{f}_0 - f_0)^2\big) + \sigma^2
\end{align}

Suppose the number of instances which are not in the training set is $m$. By Monte Carlo approximation of the expectation terms \cite{ghojogh2020sampling}, we have:
\begin{align}\label{equation_MSE_test_instance_2}
&\frac{1}{m} \sum_{i=1}^m (\widehat{f}_i - y_i)^2 = \frac{1}{m} \sum_{i=1}^m (\widehat{f}_i - f_i)^2 + \sigma^2 \implies \nonumber \\
&\sum_{i=1}^m (\widehat{f}_i - y_i)^2 = \sum_{i=1}^m (\widehat{f}_i - f_i)^2 + m \sigma^2.
\end{align}
The term $\sum_{i=1}^m (\widehat{f}_i - y_i)^2$ is the error between the predicted output and the label in the dataset. So, it is the \textit{empirical error}, denoted by \textbf{err}.
The term $\sum_{i=1}^m (\widehat{f}_i - f_i)^2$ is the error between the predicted output and true unknown label. This error is referred to as \textit{true error}, denoted by \textbf{Err}.
Therefore:
\begin{align}\label{equation_MSE_test_instance}
\textbf{err} = \textbf{Err} + m\, \sigma^2 \implies \textbf{Err} = \textbf{err} - m\, \sigma^2.
\end{align}
The term $m\, \sigma^2$ is a constant and can be ignored. Hence, in this case, the empirical error is a good estimation of the true error. Thus, we can minimize the empirical error in order to properly minimize the true error.

\subsection{Case II: Instance in the Training Set}

In case II, the instance is in the training set. For this case, we need to use a mathematical formula named SURE, introduced in the following.
Consider a multivariate random variable $\mathbb{R}^d \ni \b{z} = [z_1, \dots, z_d]^\top$ whose components are independent random variables with normal distribution, i.e., $z_i \sim \mathcal{N}(\mu_i, \sigma)$. Take $\mathbb{R}^d \ni \b{\mu} = [\mu_1, \dots, \mu_d]^\top$ and let $\mathbb{R}^d \ni \b{g}(\b{z}) = [g_1, \dots, g_d]^\top$ be a function of the random variable $\b{z}$ with $\b{g}(\b{z}): \mathbb{R}^d \rightarrow \mathbb{R}^d$.
There exists a lemma, named \textit{Stein's Lemma}, which states:
\begin{align}\label{equation_stien_lemma_vector}
\mathbb{E}\big((\b{z} - \b{\mu})^\top\, \b{g}(\b{z})\big) = \sigma^2\, \sum_{i=1}^d \mathbb{E}\big(\frac{\partial g_i}{\partial z_i}\big),
\end{align}
which is used in \textit{Stein's Unbiased Risk Estimate (SURE)} \cite{stein1981estimation}. See Appendix \ref{section_stein_proof} in this tutorial paper for the proof of Eq. (\ref{equation_stien_lemma_vector}).

If the random variable is a univariate variable, the Stein's lemma becomes:
\begin{align}\label{equation_stien_lemma_scalar}
\mathbb{E}\big((z - \mu)\, g(z)\big) = \sigma^2\, \mathbb{E}\big(\frac{\partial g(z)}{\partial z}\big).
\end{align}

Suppose we take $\varepsilon_0$, $0$, and $\widehat{f}_0 - f_0$ as the $z$, $\mu$, and $g(z)$, respectively, in the Stein's lemma for univariate variable.
Using Eq. (\ref{equation_stien_lemma_scalar}), the last term in Eq. (\ref{equation_MSE_y0}) is:
\begin{align*}
\mathbb{E}\big((\varepsilon_0 - 0) &(\widehat{f}_0 - f_0) \big) = \sigma^2\, \mathbb{E}\big(\frac{\partial (\widehat{f}_0 - f_0)}{\partial \varepsilon_0}\big) \\
&= \sigma^2\, \mathbb{E}\big(\frac{\partial \widehat{f}_0}{\partial \varepsilon_0} - \frac{\partial f_0}{\partial \varepsilon_0}\big) \overset{(a)}{=} \sigma^2\, \mathbb{E}\big(\frac{\partial \widehat{f}_0}{\partial \varepsilon_0}\big) \\ 
&\overset{(b)}{=} \sigma^2\, \mathbb{E}\big(\frac{\partial \widehat{f}_0}{\partial y_0} \times \frac{\partial y_0}{\partial \varepsilon_0}\big) \overset{(c)}{=} \sigma^2\, \mathbb{E}\big(\frac{\partial \widehat{f}_0}{\partial y_0} \big),
\end{align*}
where $(a)$ is because the true model $f$ is not dependent on the noise, $(b)$ is because of the chain rule in derivative, and $(c)$ is because:
\begin{align*}
y_0 \overset{(\ref{equation_y0})}{=} f_0 + \varepsilon_0 \implies \frac{\partial y_0}{\partial \varepsilon_0} = 1.
\end{align*}
Therefore, in this case, the Eq. (\ref{equation_MSE_y0}) is:
\begin{align}
\mathbb{E}\big(&(\widehat{f}_0 - y_0)^2 \big) = \mathbb{E}\big((\widehat{f}_0 - f_0)^2\big) + \sigma^2 - 2\sigma^2\mathbb{E}\big(\frac{\partial \widehat{f}_0}{\partial y_0} \big).
\end{align}
Suppose the number of training instances is $n$. By Monte Carlo approximation of the expectation terms \cite{ghojogh2020sampling}, we have:
\begin{align}
&\frac{1}{n} \sum_{i=1}^n (\widehat{f}_i - y_i)^2 = \frac{1}{n} \sum_{i=1}^n (\widehat{f}_i - f_i)^2 + \sigma^2 - 2\sigma^2 \frac{1}{n} \sum_{i=1}^n \frac{\partial \widehat{f}_i}{\partial y_i} 
\nonumber \\
&\implies \nonumber \\
&\sum_{i=1}^n (\widehat{f}_i - y_i)^2 = \sum_{i=1}^n (\widehat{f}_i - f_i)^2 + n \sigma^2 - 2\sigma^2 \sum_{i=1}^n \frac{\partial \widehat{f}_i}{\partial y_i}. 
\end{align}
The term $\sum_{i=1}^m (\widehat{f}_i - y_i)^2$ is the error between the predicted output and the label in the dataset. So, it is the \textit{empirical error}, denoted by \textbf{err}.
The term $\sum_{i=1}^m (\widehat{f}_i - f_i)^2$ is the error between the predicted output and true unknown label. This error is referred to as \textit{true error}, denoted by \textbf{Err}.
Therefore:
\begin{align}\label{equation_MSE_train_instance}
\textbf{Err} = \textbf{err} - n\, \sigma^2 + 2\,\sigma^2 \sum_{i=1}^n \frac{\partial \widehat{f}_i}{\partial y_i}.
\end{align}

\begin{figure}[!t]
\centering
\includegraphics[width=2.5in]{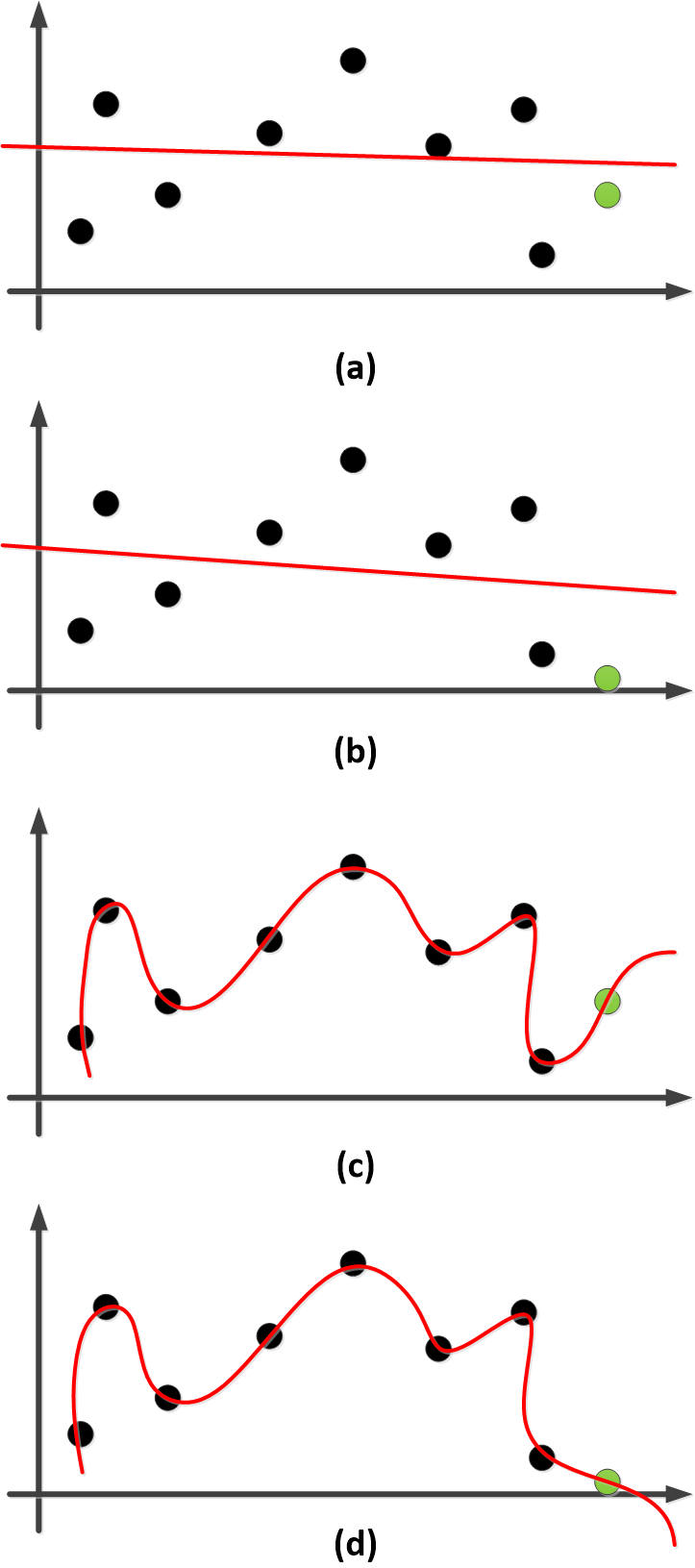}
\caption{An example for a simple and a complex model: (a) a simple model, (b) the modified simple model after moving a training instance (shown in green), (c) a complex model, and (d) the modified simple model after moving a training instance (shown in green). The complex model is impacted significantly by moving the instance while the simple model is not that much affected.}
\label{figure_overfitting_points}
\end{figure}

The last term in Eq. (\ref{equation_MSE_train_instance}) is a measure of \textit{complexity} (or \textit{overfitting}) of the model. Note that $\partial \widehat{f}_i/\partial y_i$ means if we move the $i$-th training instance, how much the model's estimation of that instance will change? This shows how much the model is complex or overfitted. For better understanding, suppose a line regressing a training set via least squares problem. If we change a point, the line will not change significantly because the model is not complex (is underfitted). On the other hand, consider a regression model passing through ``all'' the points. If we move a training point, the regressing curve changes noticeably which is because the model is very complex (overfitted). See Fig. \ref{figure_overfitting_points} illustrating the explained examples.

According to Eq. (\ref{equation_MSE_train_instance}), in the case where the instance is in the training set, the empirical error is not a good estimation of the true error. The reason is that minimization of \textbf{err} usually increases the complexity of the model, cancelling out the minimization of \textbf{Err} after some level of training.

\subsection{Estimation of $\sigma$ in MSE of the Model}

The MSE of the model in the both mentioned cases include $\sigma$ which is the standard deviation of the noise. An unbiased estimation of the variance of the noise is:
\begin{align}\label{equation_MSE_sigma_estimation}
\sigma^2 \approx \frac{1}{n-1} \sum_{i=1}^n (y_i - \widehat{f}_i),
\end{align}
which uses the training observations and their estimation by the model. However, the model's estimations of the training observations are themselves dependent on the complexity of the model. 

For the explained problem, in practice, we use an estimator with high bias and low variance in order to estimate the $\sigma$ in order not to have an estimation dependent on the complexity fo the model. For example, we use a line fitted to the training data using least squares problem (linear regression) in order to have estimations $\widehat{f}_i$'s of the $y_i$'s and then we use Eq. (\ref{equation_MSE_sigma_estimation}). Thereafter, for the sake of simplicity, we do not change the $\sigma$ (assume it is fixed) when we change the complexity of the model.

\section{Overfitting, Underfitting, and Generalization}\label{section_overfitting}

If the model is trained in an extremely simple way so that its estimation has low variance but high bias, we have \textit{underfitting}. Note that underfitting is also referred to as \textit{over-generalization}.
On the other hand, if the model is trained in an extremely complex way so that its estimation has high variance but low bias, we have \textit{overfitting}. To summarize:
\begin{itemize}
\setlength{\parskip}{0pt}
\setlength{\itemsep}{0pt plus 1pt}
\item in underfitting: low variance, high bias, and low complexity.
\item in overfitting: high variance, low bias, high complexity.
\end{itemize}

\begin{figure}[!t]
\centering
\includegraphics[width=2.2in]{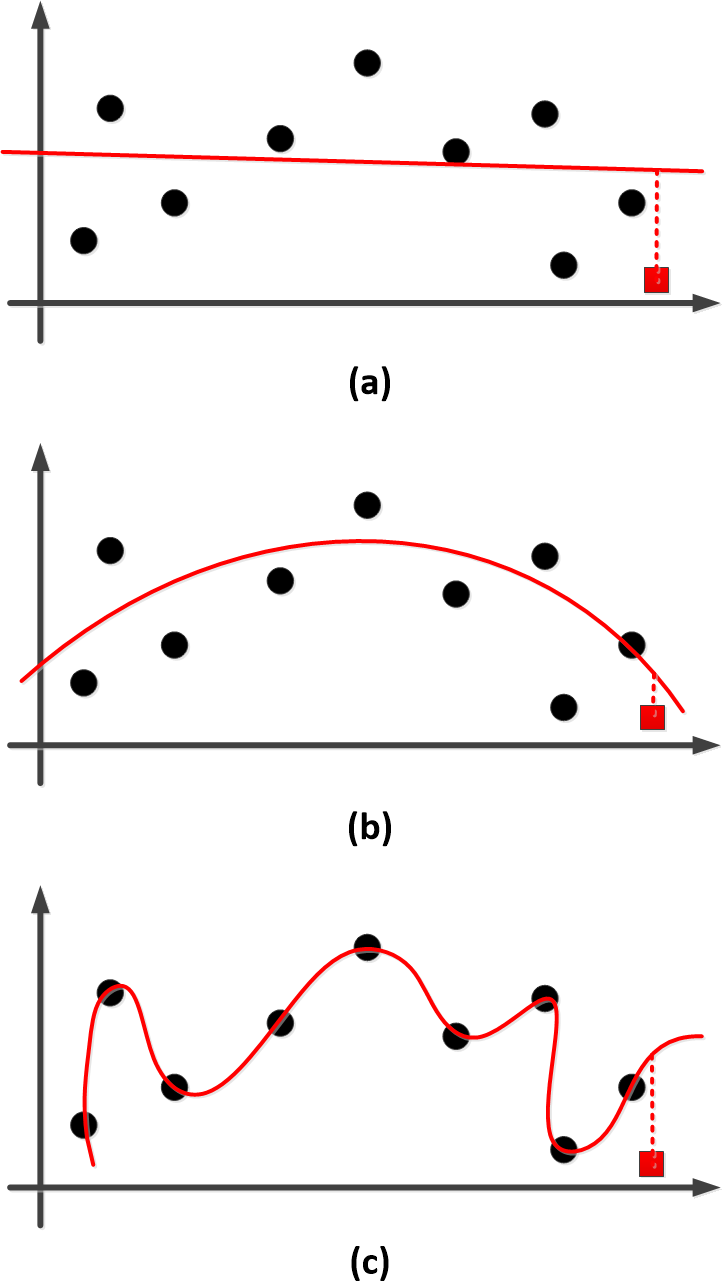}
\caption{An example for (a) underfitting, (b) good fit, and (c) overfitting. The black circles and red square are training and test instances, respectively. The red curve is the fitted curve.}
\label{figure_overfitting_example}
\end{figure}

An example for underfitting, good fit, and overfitting is illustrated in Fig. \ref{figure_overfitting_example}. As this figure shows, in both underfitting and overfitting, the estimation of a test instance might be very weak while in a good fit, the test instance, which was not seen in the training phase, is estimated well enough with smaller error. The ability of the model to estimate the unseen test (out-of-sample) data is referred to as \textit{generalization}. The lack of generalization is the reason why both overfitting and underfitting, especially overfitting, is not acceptable. In overfitting, the training error, i.e., \textbf{err}, is very small while the test (true) error, i.e., \textbf{Err}, is usually awful! 

\section{Cross Validation}\label{section_crossValidation}

\subsection{Definition}

In order to either (I) find out until which complexity we should train the model or (II) tune the parameters of the model, we should use \textit{cross validation} \cite{arlot2010survey}.
In cross validation, we divide the dataset $\mathcal{D}$ into two partitions, i.e., training set denoted by $\mathcal{T}$ and test set denoted by $\mathcal{R}$ where the union of these two subsets is the whole dataset and the intersection of them is the empty set:
\begin{align}
\mathcal{T} \cup \mathcal{R} = \mathcal{D}, \\
\mathcal{T} \cap \mathcal{R} = \varnothing.
\end{align}
The $\mathcal{T}$ is used for training the model. After the model is trained, the $\mathcal{R}$ is used for testing the performance of the model. 

We have different methods for cross validation. Two of the most well-known methods for cross validation are $K$-fold cross validation and Leave-One-Out Cross Validation (LOOCV). 

In $K$-fold cross validation, we randomly split the dataset $\mathcal{D}$ into $K$ partitions $\{\mathcal{D}_1, \dots, \mathcal{D}_K\}$ where:
\begin{align}
& |\mathcal{D}_1| \approx |\mathcal{D}_2| \approx \dots \approx |\mathcal{D}_K|, \\
& \bigcup_{i=1}^K \mathcal{D}_i = \mathcal{D}, \\
& \mathcal{D}_i \cap \mathcal{D}_j = \varnothing, ~~~ \forall i,j \in \{1, \dots, K\}, ~ i \neq j,
\end{align}
where $|.|$ denoted the cardinality of set. Sometimes, the dataset $\mathcal{D}$ is shuffled before the cross validation for better randomization. Moreover, both simple random sampling without replacement and stratified sampling \cite{barnett1974elements,ghojogh2020sampling} can be used for this splitting. 
The $K$-fold cross validation includes $K$ iterations, where in each of them, one of the partitions is used as the test set and the rest of data is used for training. The overall estimation error is the average test error of iterations.
Note that we usually have $K = 2,5,10$ in the literature but $K=10$ is the most common.
The algorithm of $K$-fold cross validation is shown in Algorithm \ref{algorithm_K_fold_CV}.

\SetAlCapSkip{0.5em}
\IncMargin{0.8em}
\begin{algorithm2e}[!t]
\DontPrintSemicolon
	Randomly split $\mathcal{D}$ into $K$ partitions with almost equal sizes.\;
	\For{$k$ from $1$ to $K$}{
	    $\mathcal{R}$ $\gets$ Partition $k$ from $\mathcal{D}$.\;
	    $\mathcal{T}$ $\gets$ $\mathcal{D} \setminus \mathcal{R}$.\;
	    Use $\mathcal{T}$ to train the model.\;
	    $\textbf{Err}_k \gets$ Use the trained model to predict $\mathcal{R}$.\; 
	}
	$\textbf{Err} \gets \frac{1}{K} \sum_{k=1}^K \textbf{Err}_k$\;
\caption{$K$-fold Cross Validation}\label{algorithm_K_fold_CV}
\end{algorithm2e}
\DecMargin{0.8em}

In LOOCV, we iterate for $|\mathcal{D}|=N$ times and in each iteration, we take one instance as the $\mathcal{R}$ (so that $|\mathcal{R}| = 1$) and the rest of instances as the training set. The overall estimation error is the average test error of iterations.
The algorithm of LOOCV is shown in Algorithm \ref{algorithm_LOOCV}.
Usually, when the size of dataset is small, LOOCV is used in order to use the most of dataset for training and then test the model properly.

If we want to train the model and then test it, the cross validation should be done using training and test sets as explained. Note that the test set and the training set should be disjoint, i.e., $\mathcal{T} \cap \mathcal{R} = \varnothing$; otherwise, we are introducing the whole or a part of the test instances to the model to learn them. Of course, in that way, the model will learn to estimate the test instances easier and better; however, in the real-world applications, the test data is not available at the time of training. Therefore, if we mistakenly have $\mathcal{T} \cap \mathcal{R} \neq \varnothing$, it is referred to as \textit{cheating} in machine learning (we call it \textit{cheating \#1} here).

\SetAlCapSkip{0.5em}
\IncMargin{0.8em}
\begin{algorithm2e}[!t]
\DontPrintSemicolon
	\For{$k$ from $1$ to $|\mathcal{D}|=N$}{
	    $\mathcal{R}$ $\gets$ Take the $k$-th instance from $\mathcal{D}$.\;
	    $\mathcal{T}$ $\gets$ $\mathcal{D} \setminus \mathcal{R}$.\;
	    Use $\mathcal{T}$ to train the model.\;
	    $\textbf{Err}_k \gets$ Use the trained model to predict $\mathcal{R}$.\; 
	}
	$\textbf{Err} \gets \frac{1}{|\mathcal{D}|} \sum_{k=1}^{|\mathcal{D}|} \textbf{Err}_k$\;
\caption{Leave-One-Out Cross Validation}\label{algorithm_LOOCV}
\end{algorithm2e}
\DecMargin{0.8em}

In some cases, the model has some parameters which need to be determined. In this case, we split the data $\mathcal{D}$ to three subsets, i.e., training set $\mathcal{T}$, test set $\mathcal{R}$, and validation set $\mathcal{V}$. Usually, we have $|\mathcal{T}| > |\mathcal{R}|$ and $|\mathcal{T}| > |\mathcal{V}|$. First, we want to find the best parameters. For this, the training set is used to train the model with different values of parameters. For every value of parameter(s), after the model is trained, it is tested on the validation set. This is performed for all desired values of parameters. The parameter value resulting in the best estimation performance on the validation set is selected to be the value of parameter(s).
After finding the values of parameters, the model is trained using the training set (where the found parameter value is used). Then, the model is tested on the test set and the estimation performance is the average test set over the cross validation iterations.

In cross validation with validation set, we have:
\begin{align}
\mathcal{T} \cap \mathcal{R} = \varnothing, ~~ \mathcal{T} \cap \mathcal{V} = \varnothing, ~~ \mathcal{V} \cap \mathcal{R} = \varnothing.
\end{align}
The validation and test sets should be disjoint because the parameters of the model should not be optimized by testing on the test set. In other words, in real-world applications, the training and validation sets are available but the test set is not available yet. If we mistakenly have $\mathcal{V} \cap \mathcal{R} \neq \varnothing$, it is referred to as \textit{cheating} in machine learning (we call it \textit{cheating \#2} here). Note that this kind of mistake is very common in the literature unfortunately, where some people optimize the parameters by testing on the test set without having a validation set.
Moreover, the training and test sets should be disjoint as explained beforehand; otherwise, that would be another kind of \textit{cheating} in machine learning (introduced before as \textit{cheating \#1}).
On the other hand, the training and validation sets should be disjoint. Although having $\mathcal{T} \cap \mathcal{V} \neq \varnothing$ is not cheating but it should not be done for the reason which will be explained later in this section.

In order to have validation set in cross validation, we usually first split the dataset $\mathcal{D}$ into $\mathcal{T}'$ and $\mathcal{R}$ where $\mathcal{T}' \cup \mathcal{R} = \mathcal{D}$ and $\mathcal{T}' \cap \mathcal{R} = \varnothing$. Then, we split the set $\mathcal{T}'$ into the training and validation sets, i.e., $\mathcal{T} \cup \mathcal{V} = \mathcal{T}'$ and $\mathcal{T} \cap \mathcal{V} = \varnothing$ and usually $|\mathcal{T}| > |\mathcal{V}|$.
The algorithms of $K$-fold cross validation and LOOCV can be modified accordingly to include the validation set. In LOOCV, we usually have $|\mathcal{V}| = 1$.

\subsection{Theory}\label{section_CV_theory}

\begin{figure}[!t]
\centering
\includegraphics[width=3.2in]{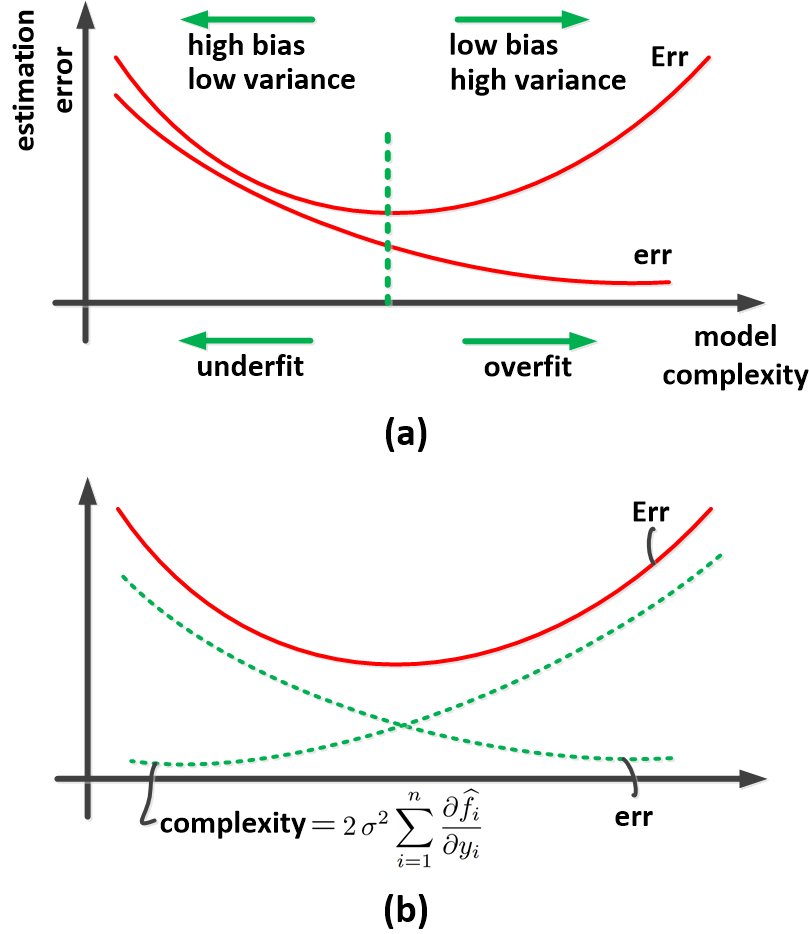}
\caption{The overfitting of model: (a) training error and true error, (b) depiction of Eq. (\ref{equation_MSE_train_instance}).}
\label{figure_overfitting_curve}
\end{figure}

Recall the Eqs. (\ref{equation_MSE_test_instance}) and (\ref{equation_MSE_train_instance}) where the true error for the test (not in the training set) and training instance are related to the training error, respectively.  
When the instance is in the training set, the true error, \textbf{Err}, and the test error, \textbf{err}, behave differently as shown in Fig. \ref{figure_overfitting_curve}-a. At the first stages of training, the \textbf{err} and \textbf{Err} both decrease; however, after some training, the model becomes more complex and goes toward overfitting. In that stage, the \textbf{Err} starts to increase. We should end the training when the \textbf{Err} starts to increase because that stage is the good fit. Usually, in order to find out when to stop training, we train the model for one stage (e.g., iteration) and then test the trained model on the validation set where the error is named \textbf{Err}. This is commonly used in training neural networks \cite{goodfellow2016deep} where \textbf{Err} is measured after every epoch for example. For neural networks, we usually save a history of \textbf{Err} for several last epochs and if the overall pattern of \textbf{Err} is increasing, we stop and take the last best trained model with least \textbf{Err}. We do this because in complex models such as neural networks, the curve of \textbf{Err} usually has some small fluctuations which we do not want to misjudge the stopping criterion based on those.  
This procedure is named \textit{early stopping} in neural networks \cite{prechelt1998early} which will be explained later in Section \ref{section_early_stopping}.

The reason why \textbf{Err} increases after a while of training is according to Eq. (\ref{equation_MSE_train_instance}). Dropping the constant $n \sigma^2$ from that expression, we have: $\textbf{Err} = \textbf{err} + 2\,\sigma^2 \sum_{i=1}^n \frac{\partial \widehat{f}_i}{\partial y_i}$ where the term $2\,\sigma^2 \sum_{i=1}^n \frac{\partial \widehat{f}_i}{\partial y_i}$ shows the model complexity. See Fig. \ref{figure_overfitting_curve}-b where both \textbf{err} and the model complexity are illustrated as a function of training stages (iterations). According to Eq. (\ref{equation_MSE_train_instance}), the \textbf{Err} is the summation of these two curves which clarifies the reason of its behavior. That is why we should not train a lot on the training set because the model will get too much fitted on the training set and will lose its ability to generalize to new unseen data.

The Fig. \ref{figure_overfitting_curve}-a and Eq. (\ref{equation_MSE_train_instance}) show that it is better to have $\mathcal{T} \cap \mathcal{V} = 0$. Otherwise, for example if we have $\mathcal{T} = \mathcal{V}$, the \textbf{Err} will be equivalent to \textbf{err} and thus it will go down even in overfitting stages. This is harmful to our training because we will not notice overfitting properly. The Eq. (\ref{equation_MSE_test_instance}) also explains that the error on validation or test set is a good measure for the true error. That is why we can use test or validation error in order to know until what stage we can train the model without overfitting.

Finally, it is noteworthy to discuss the intersections of training, test, and validation sets according to above explanations and the previous sub-section.
If we have only training and test sets without validation set:
\begin{itemize}
\item $\mathcal{T} \cap \mathcal{R} \neq \varnothing \implies $ cheating \#1
\end{itemize}
If we have training, test, and validation sets:
\begin{itemize}
\item $\mathcal{T} \cap \mathcal{R} \neq \varnothing \implies $ cheating \#1
\item $\mathcal{V} \cap \mathcal{R} \neq \varnothing \implies $ cheating \#2
\item $\mathcal{T} \cap \mathcal{V} \neq \varnothing \implies $ harmful to training (not noticing overfitting properly)
\end{itemize}
The first two items are advantageous to the model's performance on test data but that is cheating and also it may be disadvantageous to future new test data. The third item is disadvantageous to the model's performance on test data because we may not find out overfitting or we may find it out late and the generalization error will become worse; therefore, it is better not to do it.

\section{Generalized Cross Validation}

In this section, we consider the model which estimates the observations $\{y_i\}_{i=1}^N$ as:
\begin{align}\label{equation_generalized_CV_model}
\widehat{\b{y}} = \b{\Gamma}\, \b{y},
\end{align}
where $\widehat{\b{y}} = [\widehat{y}_1, \dots, \widehat{y}_N]^\top$ and $\b{y} = [y_1, \dots, y_N]^\top$ assuming that the observations for the whole dataset are available. The $\b{\Gamma} \in \mathbb{R}^{N \times N}$ is called the hat matrix because it puts a hat on $\b{y}$. 
An example of $\b{\Gamma}$ is $\b{\Gamma} = \b{X}(\b{X}^\top \b{X})^{-1} \b{X}^\top$ which is used in linear regression \cite{friedman2001elements}.
If $\gamma_{ij}$ denotes the $(i,j)$-th element of $\b{\Gamma}$, the $i$-th element of $\widehat{\b{y}}$ can be stated as:
\begin{align}
\widehat{y}_i = \sum_{j=1}^N \gamma_{ij}\, y_j.
\end{align}
Now assume that we remove the $i$-th instance for the sake of having LOOCV. Assume that $\widehat{y}_i^{(-i)}$ denotes the model's estimate of $y_i$ where the model is trained using $\mathcal{D} \setminus \{x_i\}$ (using the entire data except the $i$-th instance). We can say:
\begin{align}\label{equation_generalized_CV_midEq_1}
\widehat{y}_i^{(-i)} = \Big(\sum_{j=1}^N \gamma_{ij}\, y_j\Big) - \gamma_{ii}\, y_i + \gamma_{ij}\, \widehat{y}_i^{(-i)},
\end{align}
which means that we remove the estimation of $y_i$ using the model trained by the whole $\mathcal{D}$ and instead we put the estimation of the model trained by $\mathcal{D} \setminus \{x_i\}$. 
Adding $y_i$ to the left- and right-hand sides of Eq. (\ref{equation_generalized_CV_midEq_1}) gives:
\begin{align}\label{equation_generalized_CV_LOOCV}
y_i - \widehat{y}_i^{(-i)} = \frac{y_i - \widehat{y}_i}{1 - \gamma_{ii}}.
\end{align}
The Eq. (\ref{equation_generalized_CV_LOOCV}) means that we can do LOOCV for the model of Eq. (\ref{equation_generalized_CV_model}) without the need of iteration over the instances. We can train the model once using the whole $\mathcal{D}$ and then use Eq. (\ref{equation_generalized_CV_LOOCV}) to find the error of every iteration of LOOCV. The overall scaled mean squared error of LOOCV, then, is:
\begin{align}\label{equation_generalized_CV_LOOCV_average}
\sum_{i=1}^N \big(y_i - \widehat{y}_i^{(-i)}\big)^2 = \sum_{i=1}^N \Big(\frac{y_i - \widehat{y}_i}{1 - \gamma_{ii}}\Big)^2.
\end{align}
Suppose that we replace the $\gamma_{ii}$ by its average:
\begin{align}\label{equation_generalized_CV_average_Gamma}
\frac{1}{N} \sum_{i=1}^N \gamma_{ii} \overset{(a)}{=} \frac{1}{N} \textbf{tr}(\b{\Gamma}) \overset{(b)}{=} \frac{p}{N},
\end{align}
where $\textbf{tr}(.)$ is the trace of matrix, $(a)$ is because trace is equivalent to summation of diagonal, and $(b)$ assumes that the trace of the hat matrix is $p$. The $p$ can be considered as a the dimensionality of the subspace if the Eq. (\ref{equation_generalized_CV_model}) is considered as a projection into a subspace.

Using Eq. (\ref{equation_generalized_CV_average_Gamma}) in Eq. (\ref{equation_generalized_CV_LOOCV_average}) gives:
\begin{align}\label{equation_generalized_CV_LOOCV_final}
\sum_{i=1}^N \big(y_i - \widehat{y}_i^{(-i)}\big)^2 = \sum_{i=1}^N \Big(\frac{y_i - \widehat{y}_i}{1 - p/N}\Big)^2.
\end{align}
The Eq. (\ref{equation_generalized_CV_LOOCV_final}) is referred to as \textit{generalized cross validation} \cite{carven1979smoothing,golub1979generalized}.
It is noteworthy that the generalized cross validation can also be related to SURE \cite{stein1981estimation} which was introduced before (see \cite{li1985stein}).

\section{Regularization}\label{section_regularization}

\subsection{Definition}

We can minimize the true error, \textbf{Err}, using optimization. According to Eq. (\ref{equation_MSE_train_instance}), we have:
\begin{align}\label{equation_regularization_optimization_Err}
\text{minimize} ~~~ \textbf{err} - n\, \sigma^2 + 2\,\sigma^2 \sum_{i=1}^n \frac{\partial \widehat{f}_i}{\partial y_i}.
\end{align}
As the term $n\, \sigma^2$ is a constant, we can drop it. Moreover, calculation of $\partial \widehat{f}_i / \partial y_i$ is usually very difficult; therefore, we usually use a penalty term in place of it where the penalty increases as the complexity of the model increases in order to imitate the behavior of $\partial \widehat{f}_i / \partial y_i$.
Therefore, the optimization can be written as a \textit{regularized optimization} problem:
\begin{align}\label{equation_regularization_optimization}
\underset{\b{x}}{\text{minimize}} ~~~ \widetilde{J}(\b{x}; \theta) := J(\b{x}; \theta) + \alpha\,\Omega(\b{x}),
\end{align}
where $\theta$ is the parameter(s) of the cost function, $J(.)$ is the objective \textbf{err} to be minimized, $\Omega(.)$ is the penalty function representing the complexity of model, $\alpha >0 $ is the regularization parameter, and $\widetilde{J}(.)$ is the \textit{regularized objective function}.

The penalty function can be different things such as $\ell_2$ norm \cite{friedman2001elements}, $\ell_1$ norm \cite{tibshirani1996regression,schmidt2005least}, $\ell_{2,1}$ norm \cite{changl21}, etc. The $\ell_1$ and $\ell_{2,1}$ norms are useful for having sparsity \cite{bach2011convex,bach2012optimization}.
The sparsity is very effective because of the \textit{``bet on sparsity''} principal: ``Use a procedure that does well in sparse problems, since no procedure does well in dense problems \cite{friedman2001elements,tibshirani2015statistical}.''
The effectiveness of the sparsity can also be explained by Occam's razor \cite{domingos1999role} stating that ``simpler solutions are more likely to be correct than complex ones'' or ``simplicity is a goal in itself''.

Note that in Eqs. (\ref{equation_regularization_optimization_Err}) and (\ref{equation_regularization_optimization}), we are minimizing the \textbf{Err} (i.e., $\widetilde{J}(\b{x}; \theta)$) and not \textbf{err} (i.e., $J(\b{x}; \theta)$). 
As discussed in Sections \ref{section_overfitting} and \ref{section_crossValidation}, minimizing \textbf{err} results in overfitting. Therefore, regularization helps avoid overfitting. 

\subsection{Theory for $\ell_2$ Norm Regularization}

In this section, we briefly explain the theory behind the $\ell_2$ norm regularization \cite{friedman2001elements}, which is:
\begin{align}\label{equation_regularization_optimization_l2}
\underset{\b{x}}{\text{minimize}} ~~~ \widetilde{J}(\b{x}; \theta) := J(\b{x}; \theta) + \frac{\alpha}{2}\, ||\b{x}||_2^2.
\end{align}
The $\ell_2$ norm regularization is also referred to as \textit{ridge regression} or \textit{Tikhonov regularization} \cite{goodfellow2016deep}.

Suppose $\b{x}^*$ is minimizer of the $J(\b{x}; \theta)$, i.e.: 
\begin{align}\label{equation_regularization_derivative_J}
\nabla J(\b{x}^*; \theta) = 0.
\end{align}
The Taylor series expansion of $J(\b{x}; \theta)$ up to the second derivative at $\b{x}^*$ gives:
\begin{align}
\widehat{J}(\b{x}; \theta) &\approx J(\b{x}^*; \theta) + \nabla J(\b{x}^*; \theta) \nonumber \\
&+ \frac{1}{2} (\b{x} - \b{x}^*)^\top \b{H} (\b{x} - \b{x}^*) \nonumber \\
&= J(\b{x}^*; \theta) + \frac{1}{2} (\b{x} - \b{x}^*)^\top \b{H} (\b{x} - \b{x}^*), \label{equation_regularization_J_hat}
\end{align}
where $\b{H} \in \mathbb{R}^{d \times d}$ is the Hessian.
Using the Taylor approximation in the cost gives us \cite{goodfellow2016deep}:
\begin{align}
&\widetilde{J}(\b{x}; \theta) = \widehat{J}(\b{x}; \theta) + \frac{\alpha}{2}\, ||\b{x}||_2^2 \nonumber \\
&= J(\b{x}^*; \theta) + \frac{1}{2} (\b{x} - \b{x}^*)^\top \b{H} (\b{x} - \b{x}^*) + \frac{\alpha}{2}\, ||\b{x}||_2^2, \nonumber \\
&\frac{\partial \widetilde{J}(\b{x}; \theta)}{\partial \b{x}} = \b{0} + \b{H} (\b{x}^{\dagger} - \b{x}^*) + \alpha\, \b{x}^{\dagger}  \overset{\text{set}}{=} \b{0}, \nonumber \\
&\implies (\b{H} + \alpha\b{I})\, \b{x}^{\dagger} = \b{H} \b{x}^* \nonumber \\
&\implies \b{x}^{\dagger} = (\b{H} + \alpha\b{I})^{-1} \b{H} \b{x}^*, \label{equation_regularization_x_dagger}
\end{align}
where $\b{x}^{\dagger}$ is the minimizer of $\widetilde{J}(\b{x}; \theta)$. Note that in calculations we take $\partial J(\b{x}^*; \theta) / \partial \b{x} = \b{0}$ because the $J(\b{x}^*; \theta)$ is a constant vector with respect to $\b{x}$.
The Eq. (\ref{equation_regularization_x_dagger}) makes sense because if $\alpha=0$, which means we do not have the regularization term, we will have $\b{x}^{\dagger} = \b{x}^*$. This means that the minimizer of $\widetilde{J}(\b{x},\theta)$ will be the same as the minimizer of $J(\b{x}; \theta)$ which is correct according to Eq. (\ref{equation_regularization_optimization_l2}) where $\alpha=0$.

If we apply eigenvalue decomposition on the Hessian matrix, we will have:
\begin{align}\label{equation_regularization_Hessian_SVD}
\b{H} = \b{U}\b{\Lambda}\b{U}^\top,
\end{align}
where $\b{U}$ and $\b{\Lambda}$ contain the eigenvectors and eigenvalues, respectively.
Using this decomposition in Eq. (\ref{equation_regularization_x_dagger}) gives us:
\begin{align}
\b{x}^{\dagger} &= (\b{U}\b{\Lambda}\b{U}^\top + \alpha\b{I})^{-1} \b{U}\b{\Lambda}\b{U}^\top \b{x}^* \nonumber \\
&\overset{(a)}{=} (\b{U}\b{\Lambda}\b{U}^\top + \b{U}\b{U}^\top\alpha\b{I})^{-1} \b{U}\b{\Lambda}\b{U}^\top \b{x}^* \nonumber \\
&\overset{(b)}{=} (\b{U}\b{\Lambda}\b{U}^\top + \b{U}\alpha\b{I}\b{U}^\top)^{-1} \b{U}\b{\Lambda}\b{U}^\top \b{x}^* \nonumber \\
&= \big(\b{U}(\b{\Lambda} + \alpha\b{I})\b{U}^\top\big)^{-1} \b{U}\b{\Lambda}\b{U}^\top \b{x}^* \nonumber \\
&\overset{(c)}{=} \b{U}(\b{\Lambda} + \alpha\b{I})^{-1} \underbrace{\b{U}^{-1} \b{U}}_{\b{I}} \b{\Lambda}\b{U}^\top \b{x}^* \nonumber \\
&= \b{U}(\b{\Lambda} + \alpha\b{I})^{-1} \b{\Lambda}\b{U}^\top \b{x}^*, \label{equation_regularization_x_dagger_2}
\end{align}
where $(a)$ and $(c)$ are because $\b{U}$ is an orthogonal matrix so we have $\b{U}^{-1} = \b{U}^\top$ which yields to $\b{U}^\top \b{U} = \b{I}$ and $\b{U} \b{U}^\top = \b{I}$ (because $\b{U}$ is not truncated). The $(b)$ is because $\alpha$ is a scalar and can move between the multiplication of matrices.
The Eq. (\ref{equation_regularization_x_dagger_2}) means that we are rotating $\b{x}^*$ by $\b{U}^\top \b{x}^*$ but before rotating it back with $\b{U} \b{U}^\top \b{x}^*$, we manipulate it with the term $(\b{\Lambda} + \alpha\b{I})^{-1} \b{\Lambda}$.

Based on Eq. (\ref{equation_regularization_x_dagger_2}), we can have the following interpretations:
\begin{itemize}
\item If $\alpha = 0$, we have:
\begin{align*}
\b{x}^{\dagger} &= \b{U}\underbrace{\b{\Lambda}^{-1} \b{\Lambda}}_{\b{I}} \b{U}^\top \b{x}^* = \b{U} \b{U}^\top \b{x}^* \\
&\overset{(a)}{=} \underbrace{\b{U} \b{U}^{-1}}_{\b{I}} \b{x}^* = \b{x}^*,
\end{align*}
where $(a)$ is because $\b{U}$ is an orthogonal matrix and $(b)$ is because $\b{U}$ is a non-truncated orthogonal matrix. This means that if we do not have the penalty term, the minimizer of $\widetilde{J}(\b{x}; \theta)$ is the minimizer of $J(\b{x}; \theta)$ as expected. In other words, we are rotating the solution $\b{x}^*$ by $\b{U}^\top$ and then rotate it back by $\b{U}$.
\item If $\alpha \neq 0$, the term $(\b{\Lambda} + \alpha\b{I})^{-1} \b{\Lambda}$ is:
\begin{align*}
(\b{\Lambda} + \alpha\b{I})^{-1} \b{\Lambda} = 
\begin{bmatrix}
    \frac{\lambda_1}{\lambda_1 + \alpha} & 0 & \dots  & 0 \\
    0 & \frac{\lambda_2}{\lambda_2 + \alpha} & \dots  & 0 \\
    \vdots & \vdots & \ddots & \vdots \\
    0 & 0 & \dots  & \frac{\lambda_d}{\lambda_d + \alpha}
\end{bmatrix},
\end{align*}
where $\b{\Lambda} = \textbf{diag}([\lambda_1, \dots, \lambda_d]^\top)$.
Therefore, for the $j$-th direction of Hessian, we have $\frac{\lambda_j}{\lambda_j + \alpha}$.
\begin{itemize}
\item If $\lambda_j \gg \alpha$, we will have $\frac{\lambda_j}{\lambda_j + \alpha} \approx 1$ so for the $j$-th direction we have $(\b{\Lambda} + \alpha\b{I})^{-1} \b{\Lambda} \approx \b{I}$; therefore, $\b{x}^\dagger \approx \b{x}^*$. This makes sense because $\lambda_j \gg \alpha$ means that the $j$-th direction of Hessian and thus the $j$-th direction of $J(\b{x}; \theta)$ is large enough to be effective. Therefore, the penalty is roughly ignored with respect to it.
\item If $\lambda_j \ll \alpha$, we will have $\frac{\lambda_j}{\lambda_j + \alpha} \approx 0$ so for the $j$-th direction we have $(\b{\Lambda} + \alpha\b{I})^{-1} \b{\Lambda} \approx \b{0}$; therefore, $\b{x}^\dagger \approx \b{0}$. This makes sense because $\lambda_j \ll \alpha$ means that the $j$-th direction of Hessian and thus the $j$-th direction of $J(\b{x}; \theta)$ is small and not effective. Therefore, the penalty shrinks that direction to almost zero.
\end{itemize}
Therefore, the $\ell_2$ norm regularization keeps the effective directions but shrinks the weak directions to close to zero.

Note that the following measure is referred to as \textit{effective number of parameters} or \textit{degree of freedom} \cite{friedman2001elements}: 
\begin{align}
\sum_{j=1}^d \frac{\lambda_j}{\lambda_j + \alpha},
\end{align}
because it counts the number of effective directions as discussed above.
Moreover, the term $\lambda_j / (\lambda_j + \alpha)$ or $(\b{\Lambda} + \alpha\b{I})^{-1} \b{\Lambda}$ is called the \textit{shrinkage factor} because it shrinks the weak directions.
\end{itemize}

\subsection{Theory for $\ell_1$ Norm Regularization}

As explained before, sparsity is very useful and effective. If $\b{x} = [x_1, \dots, x_d]^\top$, for having sparsity, we should use \textit{subset selection} for the regularization:
\begin{align}\label{equation_regularization_optimization_l0}
\underset{\b{x}}{\text{minimize}} ~~~ \widetilde{J}(\b{x}; \theta) := J(\b{x}; \theta) + \alpha\, ||\b{x}||_0,
\end{align}
where:
\begin{align}
||\b{x}||_0 := \sum_{j=1}^d \mathbb{I}(x_j \neq 0) = 
\left\{
    \begin{array}{ll}
      0 & \text{if } x_j = 0, \\
      1 & \text{if } x_j \neq 0,
    \end{array}
\right.
\end{align}
is ``$\ell_0$'' norm, which is not a norm (so we used ``.'' for it) because it does not satisfy the norm properties \cite{boyd2004convex}. The ``$\ell_0$'' norm counts the number of non-zero elements so when we penalize it, it means that we want to have sparser solutions with many zero entries.
According to \cite{donoho2006most}, the convex relaxation of ``$\ell_0$'' norm (subset selection) is $\ell_1$ norm. Therefore, we write the regularized optimization as:
\begin{align}\label{equation_regularization_optimization_l1}
\underset{\b{x}}{\text{minimize}} ~~~ \widetilde{J}(\b{x}; \theta) := J(\b{x}; \theta) + \alpha\, ||\b{x}||_1.
\end{align}
Note that the $\ell_1$ regularization is also referred to as \textit{lasso} (least absolute shrinkage and selection operator) regularization \cite{tibshirani1996regression}.
Different methods exist for solving optimization having $\ell_1$ norm, such as proximal algorithm using soft thresholding \cite{parikh2014proximal} and coordinate descent \cite{wright2015coordinate,wu2008coordinate}.
Here, we explain solving the optimization using the coordinate descent algorithm.

The idea of coordinate descent algorithm is similar to the idea of Gibbs sampling \cite{casella1992explaining} where we work on the dimensions of the variable one by one.
Similar to what we did for obtaining Eq. (\ref{equation_regularization_x_dagger}), we have:
\begin{align*}
&\widetilde{J}(\b{x}; \theta) = \widehat{J}(\b{x}; \theta) + \alpha\, ||\b{x}||_1 \\
&= J(\b{x}^*; \theta) + \frac{1}{2} (\b{x} - \b{x}^*)^\top \b{H} (\b{x} - \b{x}^*) + \alpha\, ||\b{x}||_1.
\end{align*}

\begin{figure}[!t]
\centering
\includegraphics[width=2in]{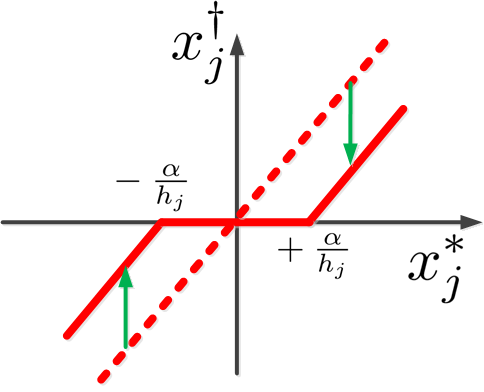}
\caption{The soft thresholding function.}
\label{figure_soft_thresholding}
\end{figure}

For simplicity in deriving an interpretable expression, we assume that the Hessian matrix is diagonal \cite{goodfellow2016deep}.
For coordinate descent, we look at the $j$-th coordinate (dimension):
\begin{align*}
&\widetilde{J}(x_j; \theta) = \widehat{J}(x_j; \theta) + \alpha\, |x_j| \\
&= J(x_j^*; \theta) + \frac{1}{2} (x_j - x_j^*)^2 h_j + \alpha\, |x_j| + c,
\end{align*}
where $\b{x} = [x_1, \dots, x_d]^\top$, $\b{x}^* = [x_1^*, \dots, x_d^*]^\top$, $h_j$ is the $(j,j)$-th element of the diagonal Hessian matrix, and $c$ is a constant term with respect to $x_j$ (not dependent to $x_j$). Taking derivative with respect to $x_j$ gives us:
\begin{align}
&\frac{\partial \widetilde{J}(x_j; \theta)}{\partial x_j} = 0 + (x_j - x_j^*)\, h_j + \alpha\, \textbf{sign}(x_j) \overset{\text{set}}{=} 0 \implies \nonumber \\
& x_j^{\dagger} = x_j^* - \frac{\alpha}{h_j}\, \textbf{sign}(x_j^{\dagger}) =
\left\{
    \begin{array}{ll}
      x_j^* - \frac{\alpha}{h_j} & \text{if } x_j^{\dagger} > 0, \\
      x_j^* + \frac{\alpha}{h_j} & \text{if } x_j^{\dagger} < 0,
    \end{array}
\right.
\end{align}
which is a soft thresholding function. This function is depicted in Fig. \ref{figure_soft_thresholding}. As can be seen in this figure, if $|x_j^*| < (\alpha / h_j)$, the solution to the regularized problem, i.e., $\b{x}_j^{\dagger}$, is zero. Recall that in $\ell_2$ norm regularization, we shrank the weak solutions close to zero; however, here in $\ell_1$ norm regularization, we are setting the weak solutions exactly to zero. That is why the solutions are relatively sparse in $\ell_1$ norm regularization. 
Notice that in $\ell_1$ norm regularization, as shown in Fig. \ref{figure_soft_thresholding}, even the strong solutions are a little shrunk (from the $x_j^{\dagger} = x_j^*$ line), the fact that we also had in $\ell_2$ norm regularization.

\begin{figure}[!t]
\centering
\includegraphics[width=2.5in]{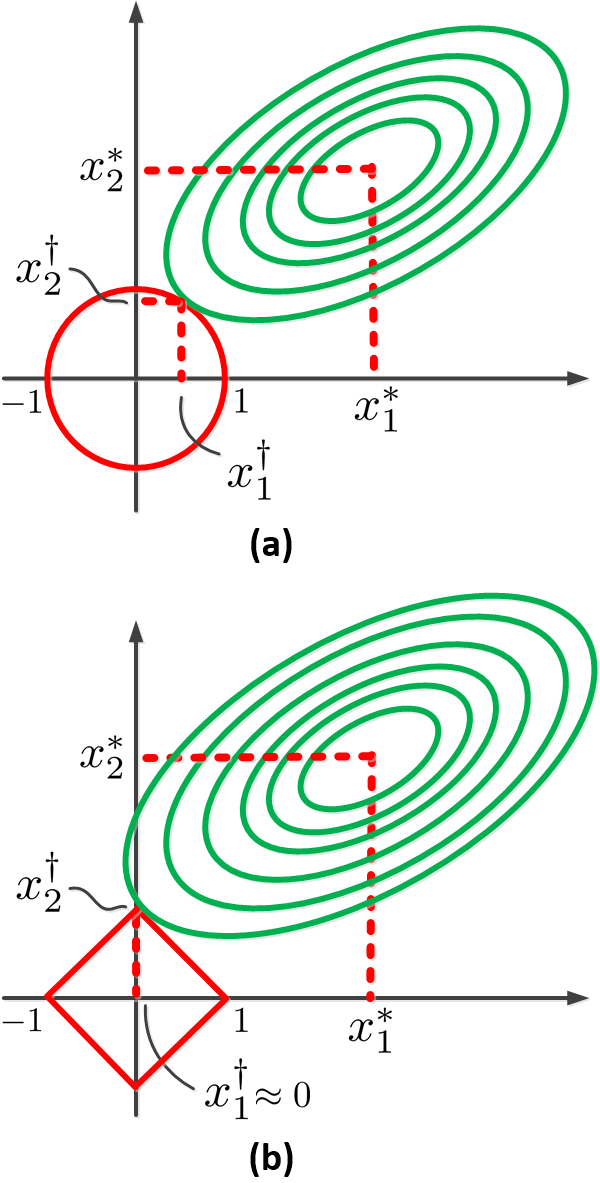}
\caption{The unit balls for $\ell_1$ and $\ell_2$ norm regularizations: (a) $\ell_2$ norm regularization and (b) $\ell_1$ norm regularization. The green curves are the contour levels of the non-regularized objective function. The red balls show the unit balls for the norm penalties. The data are assumed to be two dimensional. A third dimension can be imagined for the value of cost function. This illustration was first proposed in \cite{tibshirani1996regression}.}
\label{figure_unit_balls}
\end{figure}

Another intuition for why the $\ell_1$ norm regularization is sparse is illustrated in Fig. \ref{figure_unit_balls} \cite{tibshirani1996regression}. As this figure shows, the objective $J(\b{x}; \theta)$ has some contour levels like a bowl (if it is convex). The regularization term is also a norm ball, which is a sphere bowl (cone) for $\ell_2$ norm and a diamond bowl (cone) for $\ell_1$ norm \cite{boyd2004convex}. 
As Fig. \ref{figure_unit_balls} shows, for $\ell_2$ norm regularization, the objective and the penalty term contact at a point where some of the coordinates might be small; however, for $\ell_1$ norm, the contact point can be at some point where some variables are exactly zero. This again shows the reason of sparsity in $\ell_1$ norm regularization.

\subsection{Examples in Machine Learning: Regression, Weight Decay, Noise Injection, and Early Stopping}

\subsubsection{Linear, Ridge, and Lasso Regression}

Let $\b{X} = [\b{1}, [\b{x}_1, \dots, \b{x}_n]^\top] \in \mathbb{R}^{n \times (d+1)}$ and $\b{\beta} \in \mathbb{R}^{d+1}$.
In \textit{linear regression}, the optimization is \cite{friedman2001elements}:
\begin{align}\label{equation_linear_regression_optimization}
\underset{\b{\beta}}{\text{minimize}} ~~~ ||\b{y} - \b{X}\b{\beta}||_2^2.
\end{align}
The result of this optimization is:
\begin{align}
\b{\beta} = (\b{X}^\top \b{X})^{-1} \b{X}^\top \b{y}.
\end{align}
We can penalize the regression coefficients using $\ell_2$ norm regularization. This is referred to as \textit{ridge regression} whose optimization is \cite{friedman2001elements}:
\begin{align}\label{equation_ridge_regression_optimization}
\underset{\b{\beta}}{\text{minimize}} ~~~ ||\b{y} - \b{\beta}\b{X}||_2^2 + \frac{\alpha}{2}\, ||\b{\beta}||_2^2.
\end{align}
The result of this optimization is:
\begin{align}
\b{\beta} = (\b{X}^\top \b{X} + \alpha \b{I})^{-1} \b{X}^\top \b{y}.
\end{align}
Note that one intuition of ridge regression is that adding $\alpha \b{I}$ strengthens the main diagonal of $\b{X}^\top \b{X}$ in order to make it full-rank and non-singular for inversion. 

We can also have $\ell_1$ norm regularization, named \textit{lasso regression} \cite{tibshirani1996regression}, which makes the coefficients sparse. The optimization of lasso regression is:
\begin{align}\label{equation_lasso_regression_optimization}
\underset{\b{\beta}}{\text{minimize}} ~~~ ||\b{y} - \b{\beta}\b{X}||_2^2 + \alpha\, ||\b{\beta}||_1,
\end{align}
which dose not have a closed form solution but an iterative solution as explained before.

\subsubsection{Weight Decay}

Recall Eq. (\ref{equation_regularization_optimization_l2}).
If we replace the objective variable $\b{x}$ with the vector of neural network weights $\b{w}$, we will have:
\begin{align}\label{equation_optimization_weight_decay}
\underset{\b{w}}{\text{minimize}} ~~~ \widetilde{J}(\b{w}; \theta) := J(\b{w}; \theta) + \frac{\alpha}{2}\, ||\b{w}||_2^2,
\end{align}
which can be the loss function optimized in a neural network \cite{goodfellow2016deep}. Penalizing the weights with regularization is referred to as \textit{weight decay} \cite{krogh1992simple,chiu1994modifying}.
This penalty prevents neural network from becoming too non-linear (complex) and thus overfitted. the reason is that according to non-linear activation functions such as hyperbolic tangent, very large weights (very positive or very negative) are in the very non-linear parts of the activation functions. although neural network should not be completely linear in order to be able to learn non-linear patterns, it should not be very non-linear as well not to be overfitted to the training data. Penalizing the weights makes the weights relatively small (where the activation functions are almost linear) in to have a balance in linearity and non-linearity.

According to Eq. (\ref{equation_regularization_x_dagger_2}), the result of Eq. (\ref{equation_optimization_weight_decay}) is: 
\begin{align}\label{equation_weight_decay_w}
\b{w}^{\dagger} = \b{U}(\b{\Lambda} + \alpha\b{I})^{-1} \b{\Lambda}\b{U}^\top \b{w}^*,
\end{align}
which has the similar interpretations as we discussed before.

\subsubsection{Noise Injection to Input in Neural Networks}

In training neural networks, it is beneficial to add noise to the input \cite{matsuoka1992noise}.
One perspective to why adding noise to input helps better training of network is \textit{data augmentation} \cite{van2001art,devries2017dataset}.
Data augmentation is useful for training deep networks because they have a huge number of weights (parameters) and if we do not introduce enough training data to them, they will overfit to the training data.

Another interpretation of noise injection to input is regularization \cite{grandvalet1997noise,goodfellow2016deep}. 
Assume that the optimization of neural network is:
\begin{align}\label{equation_noise_injection_notNoise}
\underset{\b{w}}{\text{minimize}} ~~~ J:= \mathbb{E}((\widehat{y}(\b{x}) - y)^2),
\end{align}
where $\b{x}$, $\widehat{y}(\b{x})$, and $y$ are the input, the estimation (output) of network, and the training label, respectively. 
We add noise $\b{\varepsilon} \sim \mathcal{N}(\b{0}, \sigma^2\b{I})$ to the input, so the objective function changes to:
\begin{align*}
\widetilde{J} &:= \mathbb{E}((\widehat{y}(\b{x} + \b{\varepsilon}) - y)^2) \\
&= \mathbb{E}(\widehat{y}^2(\b{x} + \b{\varepsilon}) - 2y\widehat{y}(\b{x} + \b{\varepsilon}) + y^2) \\
&= \mathbb{E}(\widehat{y}^2(\b{x} + \b{\varepsilon})) - 2\mathbb{E}(y\widehat{y}(\b{x} + \b{\varepsilon})) + \mathbb{E}(y^2).
\end{align*}
Assuming that the variance of noise is small, the Taylor series expansion of $\widehat{y}(\b{x} + \b{\varepsilon})$ is:
\begin{align*}
\widehat{y}(\b{x} + \b{\varepsilon}) =\, &\widehat{y}(\b{x}) + \b{\varepsilon}^\top \nabla_{\b{x}} \widehat{y}(\b{x}) \\
&+ \frac{1}{2} \b{\varepsilon}^\top \nabla^2_{\b{x}} \widehat{y}(\b{x})\, \b{\varepsilon} + o(\b{\varepsilon}^3).
\end{align*}
Therefore:
\begin{align*}
\widetilde{J} &\approx \mathbb{E}\Big( \big(\widehat{y}(\b{x}) + \b{\varepsilon}^\top \nabla_{\b{x}} \widehat{y}(\b{x}) + \frac{1}{2} \b{\varepsilon}^\top \nabla^2_{\b{x}} \widehat{y}(\b{x})\, \b{\varepsilon} \big)^2 \Big) \\
&- 2\mathbb{E}\Big( y\widehat{y}(\b{x}) + y\b{\varepsilon}^\top \nabla_{\b{x}} \widehat{y}(\b{x}) + \frac{1}{2} y \b{\varepsilon}^\top \nabla^2_{\b{x}} \widehat{y}(\b{x})\, \b{\varepsilon} \Big) \\
&+ \mathbb{E}(y^2) \\
&= \mathbb{E}\Big(\widehat{y}(\b{x})^2 + y^2 - 2y\widehat{y}(\b{x})\Big) -2\mathbb{E}\Big(\frac{1}{2} y \b{\varepsilon}^\top \nabla^2_{\b{x}} \widehat{y}(\b{x})\, \b{\varepsilon}\Big) \\
&+ \mathbb{E}\Big(\widehat{y}(\b{x}) \b{\varepsilon}^\top \nabla_{\b{x}}^2 \widehat{y}(\b{x}) \b{\varepsilon} + (\b{\varepsilon}^\top \nabla_{\b{x}} \widehat{y}(\b{x}))^2 + o(\b{\varepsilon}^3)\Big).
\end{align*}
The first term, $\mathbb{E}(\widehat{y}(\b{x})^2 + y^2 - 2y\widehat{y}(\b{x})) = \mathbb{E}((\widehat{y}(\b{x}) - y)^2)$, is the loss function before adding the noise to the input, according to Eq. (\ref{equation_noise_injection_notNoise}). 
Also, because of $\b{\varepsilon} \sim \mathcal{N}(\b{0}, \sigma^2\b{I})$, we have $\mathbb{E}(\b{\varepsilon}^\top \b{\varepsilon}) = \sigma^2$. As the noise and the input are independent, the following term is simplified as:
\begin{align*}
\mathbb{E}\Big( (\b{\varepsilon}^\top \nabla_{\b{x}} \widehat{y}(\b{x}))^2 \Big) &\overset{\indep}{=} \mathbb{E}(\b{\varepsilon}^\top \b{\varepsilon})\, \mathbb{E}(||\nabla_{\b{x}} \widehat{y}(\b{x})||_2^2) \\
&= \sigma^2\, \mathbb{E}(||\nabla_{\b{x}} \widehat{y}(\b{x})||_2^2),
\end{align*}
and the rest of expression is simplified as:
\begin{align*}
&\mathbb{E}\Big(\widehat{y}(\b{x}) \b{\varepsilon}^\top \nabla_{\b{x}}^2 \widehat{y}(\b{x}) \b{\varepsilon} \Big) -2\mathbb{E}\Big(\frac{1}{2} y \b{\varepsilon}^\top \nabla^2_{\b{x}} \widehat{y}(\b{x})\, \b{\varepsilon}\Big) \\
&= \mathbb{E}\Big(\widehat{y}(\b{x}) \b{\varepsilon}^\top \nabla_{\b{x}}^2 \widehat{y}(\b{x}) \b{\varepsilon} \Big) -\mathbb{E}\Big(y \b{\varepsilon}^\top \nabla^2_{\b{x}} \widehat{y}(\b{x})\, \b{\varepsilon}\Big) \\
&\overset{\indep}{=} \mathbb{E}(\b{\varepsilon}^\top \b{\varepsilon})\, \mathbb{E}\big(\widehat{y}(\b{x}) \nabla_{\b{x}}^2 \widehat{y}(\b{x}) \big) - \mathbb{E}(\b{\varepsilon}^\top \b{\varepsilon})\, \mathbb{E}\big(y \nabla^2_{\b{x}} \widehat{y}(\b{x})\big) \\
&= \sigma^2\, \mathbb{E}\Big( \big(\widehat{y}(\b{x}) - y\big) \nabla_{\b{x}}^2 \widehat{y}(\b{x}) \Big).
\end{align*}
Hence, the overall loss function after noise injection to the input is simplified to:
\begin{equation}
\begin{aligned}
\widetilde{J} \approx J &+ \sigma^2\, \mathbb{E}\Big( \big(\widehat{y}(\b{x}) - y\big) \nabla_{\b{x}}^2 \widehat{y}(\b{x}) \Big) \\
&+ \sigma^2\, \mathbb{E}(||\nabla_{\b{x}} \widehat{y}(\b{x})||_2^2),
\end{aligned}
\end{equation}
which is a regularized optimization problem with $\ell_2$ norm penalty (see Eq. (\ref{equation_regularization_optimization_l2})). The penalty is on the second derivatives of outputs of neural network. This means that we do not want to have significant changes in the output of neural network. This penalization prevents from overfitting.

Note that the technique of adding noise to the input is also used in denoising autoencoders \cite{vincent2008extracting}. Moreover, an overcomplete autoencoder with one hidden layer \cite{goodfellow2016deep} (where the number of hidden neurons is greater than the dimension of data) needs a noisy input; otherwise, the mapping in the autoencoder will be just coping the input to output without learning a latent space.

It is also noteworthy that injecting noise to the weights of neural network \cite{goodfellow2016deep,ho2008weight} can be interpreted similar to injecting noise to the input. Therefore, noise injection to the weights can also be interpreted as regularization where the regularization penalty term is $\sigma^2\, \mathbb{E}(||\nabla_{\b{w}} \widehat{y}(\b{x})||_2^2)$ where $\b{w}$ is the vector of weights \cite{goodfellow2016deep}.

\subsubsection{Early Stopping in Neural Networks}\label{section_early_stopping}

As we mentioned in the explanations of Fig. \ref{figure_overfitting_curve}-a, we train neural network up to a point where the overfitting is starting. This is referred to as \textit{early stopping} \cite{prechelt1998early,yao2007early} which helps avoid overfitting \cite{caruana2001overfitting}.

According to Eq. (\ref{equation_regularization_J_hat}), we have:
\begin{align*}
\nabla_{\b{w}} \widehat{J}(\b{w}) &\approx \nabla_{\b{w}} J(\b{w}^*) + \b{H} (\b{w} - \b{w}^*) \overset{(\ref{equation_regularization_derivative_J})}{=} \b{H} (\b{w} - \b{w}^*).
\end{align*}
The gradient descent (with $\eta$ as the learning rate) used in back-propagation of neural network is \cite{boyd2004convex}: 
\begin{align*}
&\b{w}^{(t)} := \b{w}^{(t-1)} - \eta \nabla_{\b{w}} \widehat{J}(\b{w}^{(t)}) \\
& ~~~~~~~~ = \b{w}^{(t-1)} - \eta \b{H} (\b{w}^{(t-1)} - \b{w}^*) \\ 
&\implies \b{w}^{(t)} - \b{w}^* = (\b{I} - \eta \b{H}) (\b{w}^{(t-1)} - \b{w}^*),
\end{align*}
where $t$ is the index of iteration.
According to Eq. (\ref{equation_regularization_Hessian_SVD}), we have:
\begin{align*}
\b{w}^{(t)} - \b{w}^* = (\b{I} - \eta\, \b{U}\b{\Lambda}\b{U}^\top) (\b{w}^{(t-1)} - \b{w}^*).
\end{align*}
Assuming the initial weights are $\b{w}^{(0)} = 0$, we have:
\begin{align*}
&\b{w}^{(1)} - \b{w}^* = -(\b{I} - \eta\, \b{U}\b{\Lambda}\b{U}^\top) \b{w}^* \\
&\implies \b{w}^{(1)} = \big( \b{I} - (\b{I} - \eta\, \b{U}\b{\Lambda}\b{U}^\top) \big) \b{w}^* \\
&\overset{(a)}{\implies} \b{w}^{(1)} = \big( \b{U}\b{U}^\top - (\b{U}\b{U}^\top - \eta\, \b{U}\b{\Lambda}\b{U}^\top) \big) \b{w}^* \\
&\implies \b{w}^{(1)} = \b{U} \big( \b{I} - (\b{I} - \eta\, \b{\Lambda}) \big) \b{U}^\top \b{w}^*,
\end{align*}
where $(a)$ is because $\b{U}$ is a non-truncated orthogonal matrix so $\b{U}\b{U}^\top = \b{I}$.
By induction, we have:
\begin{align}
&\b{w}^{(t)} = \b{U} \big( \b{I} - (\b{I} - \eta\, \b{\Lambda})^t \big) \b{U}^\top \b{w}^*, \nonumber \\
&\implies \b{U}^\top \b{w}^{(t)} = \b{U}^\top \b{U} \big( \b{I} - (\b{I} - \eta\, \b{\Lambda})^t \big) \b{U}^\top \b{w}^*, \nonumber \\
&\overset{(a)}{\implies} \b{U}^\top \b{w}^{(t)} = \big( \b{I} - (\b{I} - \eta\, \b{\Lambda})^t \big) \b{U}^\top \b{w}^*, \label{equation_early_stopping_U_transpose_w} 
\end{align}
where $(a)$ is because $\b{U}$ is an orthogonal matrix so $\b{U}^\top\b{U} = \b{I}$.

On the other hand, recall Eq. (\ref{equation_weight_decay_w}):
\begin{align}
&\b{w}^{\dagger} = \b{U}(\b{\Lambda} + \alpha\b{I})^{-1} \b{\Lambda}\b{U}^\top \b{w}^*, \nonumber \\
&\implies \b{U}^\top \b{w}^{\dagger} = (\b{\Lambda} + \alpha\b{I})^{-1} \b{\Lambda}\b{U}^\top \b{w}^*, \nonumber \\
&\overset{(a)}{\implies} \b{U}^\top \b{w}^{\dagger} = \big( \b{I} - (\b{\Lambda} + \alpha\b{I})^{-1} \alpha \big) \b{U}^\top \b{w}^*, \label{equation_weight_decay_U_transpose_w} 
\end{align}
where $(a)$ is because of an expression rearrangement asserted in \cite{goodfellow2016deep}.
Comparing Eqs. (\ref{equation_early_stopping_U_transpose_w}) and (\ref{equation_weight_decay_U_transpose_w}) shows that early stopping can be seen as a $\ell_2$ norm regularization or weight decay \cite{goodfellow2016deep}.

Actually, the Eqs. (\ref{equation_early_stopping_U_transpose_w}) and (\ref{equation_weight_decay_U_transpose_w}) are equivalent if:
\begin{align}\label{equation_early_stopping_equating_expressions}
(\b{I} - \eta\, \b{\Lambda})^t = (\b{\Lambda} + \alpha\b{I})^{-1} \alpha,
\end{align}
for some $\eta$, $t$, and $\alpha$.
If we take the logarithm from these expressions and use Taylor series expansion for $\log(1+x)$, we have:
\begin{align}
\log (\b{I} - \eta\, \b{\Lambda})^t &= t \log (\b{I} - \eta\, \b{\Lambda}) \nonumber \\
&\approx -t\, (\eta \b{\Lambda} + \frac{1}{2} \eta^2 \b{\Lambda}^2 + \frac{1}{3} \eta^3 \b{\Lambda}^3 + \cdots), \label{equation_early_stopping_log_1}
\end{align}
\begin{align}
\log (\b{\Lambda} + \alpha\b{I})^{-1} \alpha &= -\log (\b{\Lambda} + \alpha\b{I}) + \log \alpha \nonumber \\
&= -\log (\alpha (\b{I} + \frac{1}{\alpha} \b{\Lambda})) + \log \alpha \nonumber \\
&= -\log \alpha - \log (\b{I} + \frac{1}{\alpha} \b{\Lambda}) + \log \alpha \nonumber \\
&\approx \frac{-1}{\alpha} \b{\Lambda} + \frac{1}{2\alpha^2} \b{\Lambda}^2 - \frac{1}{3\alpha^3} \b{\Lambda}^3 + \cdots. \label{equation_early_stopping_log_2}
\end{align}
Equating Eqs. (\ref{equation_early_stopping_log_1}) and (\ref{equation_early_stopping_log_2}) because of Eq. (\ref{equation_early_stopping_equating_expressions}) gives us:
\begin{align}
\alpha \approx \frac{1}{t\, \eta}, ~~~ t \approx \frac{1}{\alpha\, \eta},
\end{align}
which shows that the inverse of number of iterations is proportional to the weight decay ($\ell_2$ norm) regularization parameter. In other words, the more training iterations we have, the less we are penalizing the weights and the more the network might get overfitted. 

Moreover, some empirical studies \cite{zur2009noise} show that noise injection and weight decay have more effectiveness than early stopping for avoiding overfitting, although early stopping has its own merits.

\section{Bagging}\label{section_bagging}

\subsection{Definition}

Bagging is short for Bootstrap AGGregatING, first proposed by \cite{breiman1996bagging}. It is a meta algorithm which can be used with any model (classifier, regression, etc). 

The definition of \textit{bootstrapping} is as follows.
Suppose we have a sample $\{\b{x}_i\}_{i=1}^n$ with size $n$ where $f(\b{x})$ is the unknown distribution of the sample, i.e., $\b{x}_i \overset{iid}{\sim} f(\b{x})$. We would like to sample from this distribution but we do not know the $f(\b{x})$. Approximating the sampling from the distribution by randomly sampling from the available sample is named bootstrapping. In bootstrapping, we use simple random sampling with replacement. The drawn sample is named \textit{bootstrap sample}.

In bagging, we draw $k$ \textit{bootstrap} samples each with some sample size. Then, we train the model $h_j$ using the $j$-th bootstrap sample, $\forall j \in \{1, \dots, k\}$. Hence, we have $k$ trained models rather than one model. Finally, we \textit{aggregate} the results of estimations of the $k$ models for an instance $\b{x}$:
\begin{align}\label{equation_bagging_f_hat}
\widehat{f}(\b{x}) = \frac{1}{k} \sum_{j=1}^k h_j(\b{x}).
\end{align}
If the model is classifier, we should probably use sign function:
\begin{align}
\widehat{f}(\b{x}) = \text{sign}\big( \frac{1}{k} \sum_{j=1}^k h_j(\b{x}) \big).
\end{align}

\subsection{Theory}\label{section_bagging_theory}

Let $e_j$ denote the error of the $j$-th model in estimation of the observation of an instance. Suppose this error is a random variable with normal distribution having mean zero, i.e., $e_j \overset{iid}{\sim} \mathcal{N}(0,s)$ where $s := \sigma^2$. We denote the covariance of estimations of two trained models using two different bootstrap samples by $c$. Therefore, we have:
\begin{align}
& \mathbb{E}(e_j^2) = s \nonumber \\
&\implies \mathbb{V}\text{ar}(e_j) = \mathbb{E}(e_j^2) - (\mathbb{E}(e_j))^2 = s - 0 = s \nonumber \\
&\implies \mathbb{V}\text{ar}(h_j(\b{x})) = s,  \label{equation_bagging_variance} \\
& \mathbb{E}(e_j\, e_{\ell}) = c \nonumber \\
&\implies \mathbb{C}\text{ov}(e_j, e_{\ell}) = \mathbb{E}(e_j\, e_{\ell}) - \mathbb{E}(e_j) \mathbb{E}(e_{\ell}) \nonumber \\
&  = c - (0 \times 0) = c \implies \mathbb{C}\text{ov}(h_j(\b{x}), h_{\ell}(\b{x})) = c, \label{equation_bagging_covariance}
\end{align}
for all $j, \ell \in \{1, \dots, k\}, j \neq \ell$.

According to Eqs. (\ref{equation_bagging_f_hat}), (\ref{equation_bagging_variance}), and (\ref{equation_bagging_covariance}), we have:
\begin{align}
&\mathbb{V}\text{ar}\big(\widehat{f}(\b{x})\big) = \frac{1}{k^2} \mathbb{V}\text{ar}\big(\sum_{j=1}^k h_j(\b{x})\big) \nonumber \\
&\overset{(\ref{equation_variance_multiple})}{=} \frac{1}{k^2} \sum_{j=1}^k \mathbb{V}\text{ar}(h_j(\b{x})) \nonumber \\
&~~~~~~~~ + \frac{1}{k^2} \sum_{j=1}^k \sum_{\ell=1, \ell\neq j}^k \mathbb{C}\text{ov}(h_j(\b{x}), h_{\ell}(\b{x})) \nonumber \\
&= \frac{1}{k^2} ks + \frac{1}{k^2} k(k-1)c = \frac{1}{k} s + \frac{k-1}{k} c.
\end{align}
The obtained expression has an interesting interpretation: If two trained models with two different bootstrap samples are very correlated, we will have $c \approx s$, thus:
\begin{align}
\lim_{c \rightarrow s} \mathbb{V}\text{ar}\big(\widehat{f}(\b{x})\big) = \frac{1}{k} s + \frac{k-1}{k} s = s,
\end{align}
and if the two trained models are very different (uncorrelated), we will have $c \approx 0$, hence:
\begin{align}
\lim_{c \rightarrow 0} \mathbb{V}\text{ar}\big(\widehat{f}(\b{x})\big) = \frac{1}{k} s + \frac{k-1}{k} 0 = \frac{1}{k} s.
\end{align}
This means that if the trained models are very correlated in bagging, there is not any difference from using only one model; however, if we have different trained models, the variance of estimation improves significantly by the factor of $k$.
This also implies that bagging never is destructive; it either is not effective or improves the estimation in terms of variance \cite{buhlmann2000explaining,breiman1996bagging}.

Figure \ref{figure_overfitting_example} shows that the more complex model usually has more variance and less bias. This trade-off is shown in Fig. \ref{figure_overfitting_curve}.
Therefore, the more variance corresponds to overfitting. As bagging helps decrease the variance of estimation, it helps prevent overfitting. Therefore, bagging is a meta algorithm useful to have less variance and not to get overfitted \cite{breiman1998arcing}.
Moreover, as also will be mentioned in Section \ref{section_boosting_generalization_error_bound}, bagging can be seen as an \textit{ensemble learning} method \cite{polikar2012ensemble} which is useful because of \textit{model averaging} \cite{hoeting1999bayesian,claeskens2008model}.

\subsection{Examples in Machine Learning: Random Forest and Dropout}

\subsubsection{Random Forest}

One of the examples of using bagging in machine learning is \textit{random forest} \cite{liaw2002classification}.
In random forest, we train different models (trees) using different bootstrap samples (subsets of the training set). However, as the trees work similarly, they will be very correlated. Foe the already explained reason, this will not have a significant improvement from using one tree. Random forest addresses this issue by also sampling from the features (dimensions) of the bootstrap sample. This makes the trained trees very different and thus results in a noticeable improvement. 

\subsubsection{Dropout}

Another example of bagging is \textit{dropout} in neural networks \cite{srivastava2014dropout}. According to dropout, in every iteration of training phase, the neurons are randomly removed with probability $p = 0.5$, i.e., we sample from a Bernoulli distribution. 
This makes the training phase as training different neural networks as we have different models in bagging.
In the test time, all the neurons are used but their output is multiplied by the $p$. This imitates the model averaging of bagging in Eq. (\ref{equation_bagging_f_hat}). That is why dropout prevents neural network from overfitting. 
Another intuition of why dropout works is making the neural network sparse which is very effective because of principal of sparsity \cite{friedman2001elements,tibshirani2015statistical} or Occam's razor \cite{domingos1999role} introduced before.

\subsection{Examples in Computer Vision: HOG and SSD}

\subsubsection{Histogram of Oriented Gradients}

An example of bagging is Histogram of Oriented Gradients (HOG) \cite{dalal2005histograms} used in computer vision, especially for human detection in images. 
In HOG, different cells or blocks are used each of which includes a histogram of gradients of a sub-region of image. Finally, using bagging, the histograms are combined into one histogram. The effectiveness of HOG is because of effectiveness of bagging.

\subsubsection{Single Shot multi-box Detector}

Single Shot multi-box Detector (SSD) \cite{liu2016ssd} is another usage of bagging in computer vision and object detection using deep neural networks \cite{lecun2015deep,goodfellow2016deep}.
In SSD, a set of bounding boxes (i.e., the models in bagging) with different sizes are used which are processed and learned using convolution layers in neural network. Some of the boxes are matched and their weighted summation is used as the loss function of the neural network to optimize. 

\section{Boosting}

\subsection{Definition}

Boosting is a meta algorithm which can be used with any model (classifier, regression, etc). For binary classification, for example, if we use boosting with a classifier even slightly better than flipping a coin, we will have a strong classifier (we will explain the reason later in Section \ref{section_boosting_generalization_error_bound}). Thus, we can say boosting makes the estimation or classification very strong. In other words, boosting addresses the question whether a strong classifier can be obtained from a set of weak classifiers \cite{kearns1988thoughts,kearns1994cryptographic}.

\begin{figure}[!t]
\centering
\includegraphics[width=2.6in]{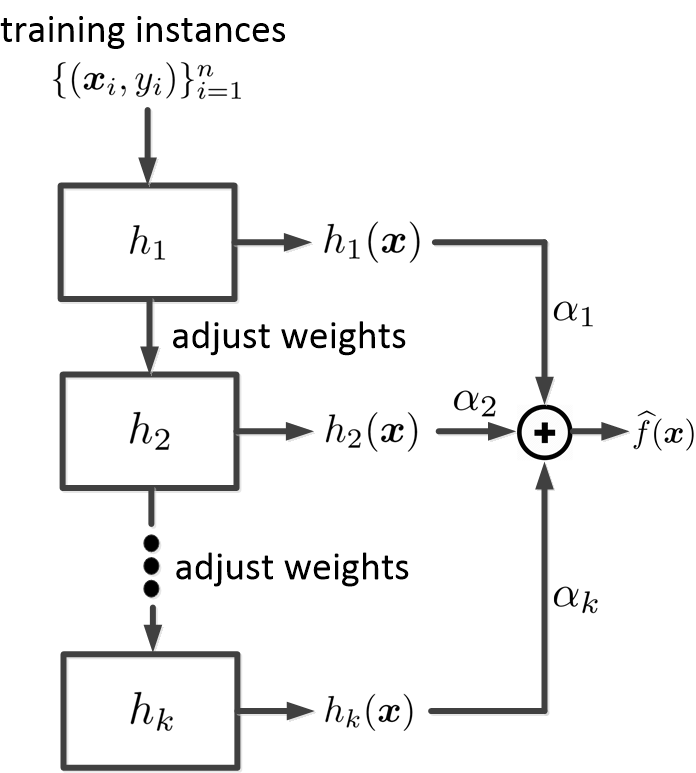}
\caption{The training phase in boosting $k$ models.}
\label{figure_boosting}
\end{figure}

The idea of boosting is to learn $k$ models in a hierarchy where every model gives more attention (larger weight) to the instances misclassified (or estimated very badly) by the previous model. Figure \ref{figure_boosting} shows this hierarchy. Finally, the overall estimation or classification is a weighted summation (average) of the $k$ estimations.
For an instance $\b{x}$, we have:
\begin{align}\label{equation_boosting_f_hat}
\widehat{f}(\b{x}) = \sum_{j=1}^k \alpha_j\, h_j(\b{x}).
\end{align}
If the model is classifier, we should probably use sign function:
\begin{align}
\widehat{f}(\b{x}) = \text{sign}\big( \sum_{j=1}^k \alpha_j h_j(\b{x}) \big),
\end{align}
which is equivalent to \textit{majority voting} among the trained classifiers.

Different methods have been proposed for boosting, one of the most well-known ones is AdaBoost (Adaptive Boosting) \cite{freund1996experiments}.
The algorithm of AdaBoost for binary classification is shown in Algorithm \ref{algorithm_AdaBoost}.
In this algorithm, $L_j$ is the cost function minimized in the $j$-th model $h_j$, the $\mathbb{I}(.)$ is the indicator function which is one and zero if its condition is and is not satisfied, respectively, and $w_i$ is the weight associated to the $i$-th instance for weighting it as the input to the next layer of boosting. 
Here, we can have several cases which help us understand the interpretation of the AdaBoost algorithm: 
\begin{itemize}
\item if an instance is correctly classified, the $\mathbb{I}(y_i \neq h_j(\b{x}_i))$ is zero and thus the $w_i$ will be still $w_i$ without any change. This makes sense because the correctly classified instance should not gain a significant weight in the next layer of boosting.
\item if an instance is misclassified, the $\mathbb{I}(y_i \neq h_j(\b{x}_i))$ is one. In this case, we can have two sub-cases:
\begin{itemize}
\item If the classifier which classified that instance was a bad classifier, its cost would be like flipping a coin, i.e., $L_j = 0.5$. Therefore, we will have $\alpha_j = \log(1)=0$ and again the $w_i$ will still be $w_i$ without any change. This makes sense because we cannot trust the bad classifier whether the instance is correctly or incorrectly classified and thus we should not make any decision based on that.
\item If the classifier which classified that instance was a good classifier, then we have $L_j = 0.5$ and as we also have $\mathbb{I}(y_i \neq h_j(\b{x}_i)) = 1$, the weight will change as $w_i := w_i\, \exp(\alpha_j)$. This also is intuitive because the previous model in the boosting was a good classifier and we can trust it and that good classifier could not classify the instance correctly. therefore, we should notice that instance more in the next model in the boosting hierarchy. 
\end{itemize}
\end{itemize}
Note that the cost in AdaBoost is:
\begin{align}\label{equation_AdaBoost_cost}
L_j = \frac{\sum_{i=1}^n w_i\, \mathbb{I}(y_i \neq h_j(\b{x}_i))}{\sum_{i=1}^n w_i},
\end{align}
which makes sense because it gets larger if the observations of more instances are estimated incorrectly.

\SetAlCapSkip{0.5em}
\IncMargin{0.8em}
\begin{algorithm2e}[!t]
\DontPrintSemicolon
	\textbf{Initialize} $w_i = 1/n, \forall i \in \{1, \dots, n\}$\;
	\For{$j$ from $1$ to $k$}{
	    $h_j(\b{x}) = \arg \min L_j$\;
	    $\alpha_j = \log(\frac{1-L_j}{L_j})$\; \label{algorithm_AdaBoost_alpha}
	    $w_i = w_i\, \exp\big(\alpha_j\, \mathbb{I}(y_i \neq h_j(\b{x}_i))\big)$\; \label{algorithm_AdaBoost_w}
	}
\caption{The AdaBoost Algorithm}\label{algorithm_AdaBoost}
\end{algorithm2e}
\DecMargin{0.8em}

\subsection{Theory Based on Additive Models}

Additive models \cite{hastie1986generalized} can be used to explain why boosting works \cite{friedman2000additive,rojas2009adaboost}. 
In additive model, we map the data as $\b{x} \mapsto \phi_j(\b{x}), \forall j\in \{1, \dots, k\}$ and then add them using some weights $\beta_j$'s:
\begin{align}\label{equation_additive_model}
\phi(\b{x}) = \sum_{j=1}^k \beta_j\, \phi_j(\b{x}).
\end{align}
A well-known example of the additive model is Radial Basis Function (RBF) neural network (here with $k$ hidden nodes) which uses Gaussian mappings \cite{broomhead1988multivariable,schwenker2001three}.

Now, consider a cost function for an instance as:
\begin{align}
L(y, h(\b{x})) := \exp(- y\, h(\b{x})),
\end{align}
where $y$ is the observation or label for $\b{x}$ and $h(\b{x})$ is the model's estimation of $y$. This cost is intuitive because when the instance is misclassified, the signs of $y$ and $h(\b{x})$ will be different and the cost will be large, while in case of correct classification, the signs are similar and the cost is small.
If we add up the cost over the $n$ training instances, we have:
\begin{align}\label{equation_boosting_L_t}
L_t(y, h(\b{x})) := \sum_{i=1}^n \exp(- y_i\, h(\b{x}_i)),
\end{align}
where $L_t$ denotes the total cost.

In Eq. (\ref{equation_additive_model}), if we rename the mapping to $h(\b{x})$, which is the model used in boosting, we will have:
\begin{align}
h(\b{x}) = \sum_{j=1}^k \beta_j\, h_j(\b{x}).
\end{align}
We can write this expression as a \textit{forward stage-wise additive model} \cite{friedman2000additive,rojas2009adaboost} in which we work with the models one by one where we add up the previously worked models:
\begin{align}
&f_{q-1}(\b{x}) = \sum_{j=1}^{q-1} \beta_j\, h_j(\b{x}), \label{equation_forward_additive_model_1}  \\
&f_{q}(\b{x}) = f_{q-1}(\b{x}) + \beta_q\, h_q(\b{x}), ~~~ q \leq k, \label{equation_forward_additive_model_2}
\end{align}
where $h(\b{x}) = f_k(\b{x})$.
Therefore, minimizing the cost, i.e., Eq. (\ref{equation_boosting_L_t}), for the $j$-th model in the additive manner is:
\begin{align*}
&\min_{\beta_j, h_j} \sum_{i=1}^n \exp\big(\! - y_i\, [f_{j-1}(\b{x}_i) + \beta_j\, h_j(\b{x}_i)]\big) \\
&= \min_{\beta_j, h_j} \sum_{i=1}^n \exp(- y_i\, f_{j-1}(\b{x}_i)) \exp(-y_i\, \beta_j\, h_j(\b{x}_i)).
\end{align*}
The first term is a constant with respect to $\beta_j$ and $h_j$ so we name it by $w_i$: 
\begin{align}\label{equation_boosting_w}
w_i := \exp(- y_i\, f_{j-1}(\b{x}_i)).
\end{align}
Thus:
\begin{align*}
&\min_{\beta_j, h_j} \sum_{i=1}^n w_i \exp(-y_i\, \beta_j\, h_j(\b{x}_i)).
\end{align*}
As in binary AdaBoost, we have $\pm 1$ for $y_i$ and $h_j$, we can say:
\begin{align*}
&\min_{\beta_j, h_j} \exp(-\beta_j) \sum_{i=1}^n w_i\, \mathbb{I}(y_i = h_j(\b{x}_i)) \\
&~~~~~~~~ + \exp(\beta_j) \sum_{i=1}^n w_i\, \mathbb{I}(y_i \neq h_j(\b{x}_i)) \\
&\overset{(a)}{=} \min_{\beta_j, h_j} \exp(-\beta_j) \sum_{i=1}^n w_i \\
&~~~~~~~~ - \exp(-\beta_j) \sum_{i=1}^n w_i\, \mathbb{I}(y_i \neq h_j(\b{x}_i)) \\
&~~~~~~~~ + \exp(\beta_j) \sum_{i=1}^n w_i\, \mathbb{I}(y_i \neq h_j(\b{x}_i)), \\
\end{align*}
where $(a)$ is because:
\begin{align*}
\sum_{i=1}^n w_i\, \mathbb{I}(y_i = h_j(\b{x}_i)) = \sum_{i=1}^n w_i - \sum_{i=1}^n w_i\, \mathbb{I}(y_i \neq h_j(\b{x}_i)).
\end{align*}
For the sake of minimization, we take the derivative:
\begin{align*}
&\frac{\partial L_t}{\partial \beta_j} = 
-\exp(-\beta_j) \sum_{i=1}^n w_i \\
&~~~~~~~~ + \exp(-\beta_j) \sum_{i=1}^n w_i\, \mathbb{I}(y_i \neq h_j(\b{x}_i)) \\
&~~~~~~~~ + \exp(\beta_j) \sum_{i=1}^n w_i\, \mathbb{I}(y_i \neq h_j(\b{x}_i)) \overset{\text{set}}{=} 0,
\end{align*}
which gives:
\begin{align}
&\implies (\exp(-\beta_j) + \exp(\beta_j)) \times \nonumber \\
&~~~~~~~~~ \frac{\mathbb{I}(y_i \neq h_j(\b{x}_i))}{\sum_{i=1}^n w_i} = \exp(-\beta_j) \nonumber \\
&\overset{(\ref{equation_AdaBoost_cost})}{\implies} (\exp(-\beta_j) + \exp(\beta_j))\, L_j = \exp(-\beta_j) \nonumber \\
&\implies L_j = \frac{\exp(-\beta_j)}{\exp(-\beta_j) + \exp(\beta_j)} \nonumber \\
&\implies \exp(2\beta_j) = \frac{1 - L_j}{L_j}  \implies 2\beta = \log(\frac{1-L_j}{L_j}) \nonumber \\
&\overset{(a)}{\implies} \alpha_j = 2\beta_j, \label{equation_boosting_alpha_beta}
\end{align}
where $(a)$ is because of the line \ref{algorithm_AdaBoost_alpha} in Algorithm \ref{algorithm_AdaBoost}.

According to Eqs. (\ref{equation_forward_additive_model_1}), (\ref{equation_forward_additive_model_2}), and (\ref{equation_boosting_w}), we have:
\begin{align}\label{equation_boosting_w_2}
w_i := w_i\, \exp(-y_i\, \beta_j\, h_j(\b{x}_i)).
\end{align}
As we have $y_i\, h_j(\b{x}_i) = \pm 1$, we can say:
\begin{align}\label{equation_boosting_y_h}
-y_i\, h_j(\b{x}_i) = 2\, \mathbb{I}(y_i \neq h_j(\b{x}_i)) - 1.
\end{align}
According Eqs. (\ref{equation_boosting_alpha_beta}), (\ref{equation_boosting_w_2}), and (\ref{equation_boosting_y_h}), we have:
\begin{align}
w_i := w_i\, \exp\big(\alpha_j\, \mathbb{I}(y_i \neq h(\b{x}_i))\big)\, \exp(-\beta_j),
\end{align}
which is equivalent to the line \ref{algorithm_AdaBoost_w} in Algorithm \ref{algorithm_AdaBoost} with a factor of $\exp(-\beta_j)$.
This factor does not have impact on whether the instance is correctly classified or not.

\subsection{Theory Based on Maximum Margin}

\subsubsection{Upper Bound on the Generalization Error of Boosting}\label{section_boosting_generalization_error_bound}

There is an upper bound on the generalization error of boosting \cite{schapire1998boosting}. 
In binary boosting, we have $\pm 1$ for $y_i$ and also the sign of $\widehat{f}(\b{x}_i)$ is important; therefore, $y_i\,\widehat{f}(\b{x}_i) < 0$ means that we have error for estimating the $i$-th instance.
Thus, for an error, we have:
\begin{align}\label{equation_boosting_y_f_theta}
y_i\,\widehat{f}(\b{x}_i) \leq \theta,
\end{align}
for a $\theta>0$. 
Recall the Eq. (\ref{equation_boosting_f_hat}). We can normalize this equation because the sign of it is important:
\begin{align}\label{equation_boosting_f_hat_normalized}
\widehat{f}(\b{x}_i) = \frac{\sum_{j=1}^k \alpha_j\, h_j(\b{x}_i)}{\sum_{j=1}^k \alpha_j}.
\end{align}
According to Eqs. (\ref{equation_boosting_y_f_theta}) and (\ref{equation_boosting_f_hat_normalized}), we have:
\begin{align*}
&y_i\,\widehat{f}(\b{x}_i) \leq \theta \Longleftrightarrow y_i \sum_{j=1}^k \alpha_j\, h_j(\b{x}_i) \leq \theta \sum_{j=1}^k \alpha_j \\
&\Longleftrightarrow \exp\big(-y_i \sum_{j=1}^k \alpha_j\, h_j(\b{x}_i) + \theta \sum_{j=1}^k \alpha_j\big) \geq 1.
\end{align*}
Therefore, in terms of probability, we have:
\begin{align}
&\mathbb{P}(y_i\,\widehat{f}(\b{x}_i) \leq \theta) \nonumber \\
&~~~= \mathbb{P}\Big(\exp\big(-y_i \sum_{j=1}^k \alpha_j\, h_j(\b{x}_i) + \theta \sum_{j=1}^k \alpha_j\big) \geq 1\Big). \label{equation_boosting_probs_equal}
\end{align}

According to the Markov's inequality which is (for $a>0$ and a random variable $X$):
\begin{align}
\mathbb{P}(X \geq a) \leq \frac{\mathbb{E}(X)}{a},
\end{align}
and Eq. (\ref{equation_boosting_probs_equal}), we have (take $a=1$ and the exponential term as $X$ in Markov's inequality):
\begin{align}
&\mathbb{P}(y_i\,\widehat{f}(\b{x}_i) \leq \theta) \nonumber \\
&\leq \mathbb{E}\Big(\exp\big(-y_i \sum_{j=1}^k \alpha_j\, h_j(\b{x}_i) + \theta \sum_{j=1}^k \alpha_j\big)\Big) \nonumber \\
&\overset{(a)}{=} \exp\big(\theta \sum_{j=1}^k \alpha_j\big)\, \mathbb{E}\Big(\exp\big(-y_i \sum_{j=1}^k \alpha_j\, h_j(\b{x}_i) \big)\Big) \nonumber \\
&\overset{(b)}{=} \frac{1}{n} \exp\big(\theta \sum_{j=1}^k \alpha_j\big)\, \sum_{i=1}^n \exp\big(-y_i \sum_{j=1}^k \alpha_j\, h_j(\b{x}_i) \big), \label{equation_boosting_bound_prob_1}
\end{align}
where $(a)$ is because the expectation is with respect to the data, i.e., $\b{x}_i$ and $y_i$ and $(b)$ is according to definition of expectation.

Recall the line \ref{algorithm_AdaBoost_w} in Algorithm \ref{algorithm_AdaBoost}:
\begin{align*}
w_i^{(j+1)} = w_i^{(j)}\, \exp\big(\alpha_j\, \mathbb{I}(y_i \neq h_j(\b{x}_i))\big),
\end{align*}
which can be restated as:
\begin{align*}
w_i^{(j+1)} = w_i^{(j)}\, \exp\big(\!-y_i\, \alpha_j\, h_j(\b{x}_i)\big),
\end{align*}
because $y_i = \pm 1$ and $h_j(\b{x}_i) = \pm 1$.
It is not harmful to AdaBoost if we use the normalized weights:
\begin{align}\label{equation_boosting_normalized_weight}
w_i^{(j+1)} = \frac{w_i^{(j)}\, \exp\big(\!-y_i\, \alpha_j\, h_j(\b{x}_i)\big)}{z_j},
\end{align}
where:
\begin{align}\label{equation_boosting_normalized_weight_denominator}
z_j := \sum_{i=1}^n w_i^{(j)} \exp\big(\!-y_i\, \alpha_j\, h_j(\b{x}_i)\big).
\end{align}
Considering that $w_i^{(1)} = 1/n$, we can have recursive expression for the weights:
\begin{align}
&w_i^{(k+1)} = \frac{w_i^{(k)}\, \exp\big(\!-y_i\, \alpha_k\, h_k(\b{x}_i)\big)}{z_k} \nonumber \\
&= w_i^{(1)} \times \frac{1}{z_k \times \dots \times z_1} \times \nonumber \\
&\exp\big(\!-y_i\, \alpha_k\, h_k(\b{x}_i)\big) \times \dots \times \exp\big(\!-y_i\, \alpha_1\, h_1(\b{x}_i)\big) \nonumber \\
&= \frac{1}{n} \times \frac{1}{\prod_{j=1}^k z_j} \times \prod_{j=1}^k \exp\big(\!-y_i\, \alpha_j\, h_j(\b{x}_i)\big) \nonumber \\
&= \frac{1}{n} \times \frac{1}{\prod_{j=1}^k z_j} \times \exp\big(\!-y_i \sum_{j=1}^k \alpha_j\, h_j(\b{x}_i)\big). \label{equation_boosting_bound_prob_2}
\end{align}
We continue the Eq. (\ref{equation_boosting_bound_prob_1}):
\begin{align*}
&\mathbb{P}(y_i\,\widehat{f}(\b{x}_i) \leq \theta) \\
&\leq \frac{1}{n} \exp\big(\theta \sum_{j=1}^k \alpha_j\big)\, \sum_{i=1}^n \exp\big(-y_i \sum_{j=1}^k \alpha_j\, h_j(\b{x}_i) \big) \\
&\overset{(\ref{equation_boosting_bound_prob_2})}{=} \exp\big(\theta \sum_{j=1}^k \alpha_j\big)\, \Big(\prod_{j=1}^k z_j\Big) \sum_{i=1}^n w_i^{(k+1)}.
\end{align*}
According to Eqs. (\ref{equation_boosting_normalized_weight}) and (\ref{equation_boosting_normalized_weight_denominator}), we have:
\begin{align*}
\sum_{i=1}^n w_i^{(j+1)} = \frac{\sum_{i=1}^n w_i^{(j)}\, \exp\big(\!-y_i\, \alpha_j\, h_j(\b{x}_i)\big)}{\sum_{i=1}^n w_i^{(j)} \exp\big(\!-y_i\, \alpha_j\, h_j(\b{x}_i)\big)} = 1.
\end{align*}
Therefore:
\begin{align}\label{equation_boosting_generalization_error_1}
\therefore ~~~ \mathbb{P}(y_i\,\widehat{f}(\b{x}_i) \leq \theta) \leq \exp\big(\theta \sum_{j=1}^k \alpha_j\big)\, \Big(\prod_{j=1}^k z_j\Big).
\end{align}

On the other hand, according to Eq. (\ref{equation_boosting_normalized_weight_denominator}), we have:
\begin{align}
z_j &= \sum_{i=1}^n w_i^{(j)} \exp\big(\!-y_i\, \alpha_j\, h_j(\b{x}_i)\big) \nonumber \\
&= \sum_{i=1}^n w_i^{(j)} \exp(-\alpha_j)\, \mathbb{I}(y_i = h_j(\b{x}_i)) \nonumber \\
&~~~~ + \sum_{i=1}^n w_i^{(j)} \exp(\alpha_j)\, \mathbb{I}(y_i \neq h_j(\b{x}_i)) \nonumber \\
&= \exp(-\alpha_j) \sum_{i=1}^n w_i^{(j)} \mathbb{I}(y_i = h_j(\b{x}_i)) \nonumber \\
&~~~~ + \exp(\alpha_j) \sum_{i=1}^n w_i^{(j)} \mathbb{I}(y_i \neq h_j(\b{x}_i)). \label{equation_boosting_z_expansion}
\end{align}
Recall Eq. (\ref{equation_boosting_normalized_weight}) for $w_i^{j+1}$. This is in the range $[0,1]$ and its summation over error cases can be considered as the probability of error:
\begin{align}
\sum_{i=1}^n w_i^{(j)} \mathbb{I}(y_i \neq h_j(\b{x}_i)) = \mathbb{P}(y_i \neq h_j(\b{x}_i)) \overset{(a)}{=} L_j,
\end{align}
where $(a)$ is because the Eq. (\ref{equation_AdaBoost_cost}) is the cost which is the probability of error.
Therefore, the Eq. (\ref{equation_boosting_z_expansion}) becomes:
\begin{align*}
z_j = \exp(-\alpha_j)\, (1 - L_j) + \exp(\alpha_j)\, L_j.
\end{align*}

Recall the $\alpha_j$ in line \ref{algorithm_AdaBoost_alpha} in Algorithm \ref{algorithm_AdaBoost}. Scaling it is not harmful to AdaBoost:
\begin{align}\label{equation_boosting_alpha_scaled}
\alpha_j = \frac{1}{2} \log(\frac{1-L_j}{L_j}).
\end{align}
Therefore, we can have:
\begin{align}\label{equation_boosting_z_relatedTo_L}
z_j = 2 \sqrt{L_j (1 - L_j)}.
\end{align}
Plugging Eqs. (\ref{equation_boosting_alpha_scaled}) and (\ref{equation_boosting_z_relatedTo_L}) in Eq. (\ref{equation_boosting_generalization_error_1}) gives:
\begin{align*}
&\mathbb{P}(y_i\,\widehat{f}(\b{x}_i) \leq \theta) \\
&\leq \exp\big(\frac{1}{2} \theta \sum_{j=1}^k \log(\frac{1-L_j}{L_j}) \big)\, \Big(2^k \prod_{j=1}^k \sqrt{L_j (1 - L_j)}\Big) \\
&= 2^k \exp\big(\sum_{j=1}^k \log((\frac{1-L_j}{L_j})^{\theta/2}) \big)\, \prod_{j=1}^k \sqrt{L_j (1 - L_j)} \\
&= 2^k \prod_{j=1}^k \exp\big(\log((\frac{1-L_j}{L_j})^{\theta/2}) \big)\, \prod_{j=1}^k \sqrt{L_j (1 - L_j)},
\end{align*}
which simplifies to the upper bound on the generalization error of AdaBoost \cite{schapire1998boosting}:
\begin{align}\label{equation_boosting_generalization_error_bound}
&\mathbb{P}\big(y_i\,\widehat{f}(\b{x}_i) \leq \theta\big) \leq 2^k \prod_{j=1}^k \sqrt{L_j^{1-\theta} (1 - L_j)^{1+\theta}},
\end{align}
where $\mathbb{P}\big(y_i\,\widehat{f}(\b{x}_i) \leq \theta\big)$ is the probability that the generalization (true) error for the $i$-th instance is less than $\theta > 0$.

According to Eq. (\ref{equation_AdaBoost_cost}), we have $L_j \in [0,1]$. If we have $L_j \leq 0.5 - \xi$, where $\xi \in (0,0.5)$, the Eq. (\ref{equation_boosting_generalization_error_bound}) becomes:
\begin{align}\label{equation_boosting_generalization_error_bound_2}
\mathbb{P}\big(y_i\,\widehat{f}(\b{x}_i) \leq \theta\big) \leq \bigg(\!\sqrt{(1-2\xi)^{1-\theta} (1+2\xi)^{1+\theta}}\bigg)^k,
\end{align}
which is a very good upper bound because if $\theta < \xi$, we have $\sqrt{(1-2\xi)^{1-\theta} (1+2\xi)^{1+\theta}} < 1$; thus, the probability of error, $\mathbb{P}\big(y_i\,\widehat{f}(\b{x}_i) \leq \theta\big)$, decreases \textit{exponentially} with $k$ which is the number of models used in boosting.  
This shows that boosting helps us reduce the generalization error and thus helps us avoid overfitting. In other words, because of the bound on generalization error, boosting overfits very hardly.

If $\xi$ is a very small positive number, the $L_j \leq 0.5 - \xi$ is a little smaller than $0.5$, i.e., $L_j \lessapprox 0.5$. As we are discussing binary classification in boosting, $L_j = 0.5$ means random classification by flipping a coin. Therefore, for having the great bound of Eq. (\ref{equation_boosting_generalization_error_bound_2}), having weak base models (a little better than random decision) suffices. This shows the effectiveness of boosting.
Note that a very small $\xi$ means a very small $\theta$ because of $\theta < \xi$; therefore, it means a very small probability of error because of $\mathbb{P}\big(y_i\,\widehat{f}(\b{x}_i) \leq \theta\big)$.

It is noteworthy that both boosting and bagging can be seen as \textit{ensemble learning} \cite{polikar2012ensemble} (or majority voting) methods which use \textit{model averaging} \cite{hoeting1999bayesian,claeskens2008model} and are very effective in learning theory.
Moreover, both boosting and bagging reduce the variance of estimation \cite{breiman1998arcing,schapire1998boosting}, especially for the models with high variance of estimation such as trees \cite{quinlan1996bagging}.

In the above, we analyzed boosting for \textit{binary} classification. A similar discussion can be done for \textit{multi-class} classification in boosting and find an upper bound on the generalization error (see the appendix in \cite{schapire1998boosting} for more details).

\subsubsection{Boosting as Maximum Margin Classifier}

In another perspective, the found upper bound for boosting shows that boosting can be seen as a method to increase (maximize) the margins of training error which results in a good generalization error \cite{boser1992training}. This phenomenon is the base for the theory of Support Vector Machines (SVM) \cite{cortes1995support,burges1998tutorial}.
In the following, we analyze the analogy between maximum margin classifier (i.e., SVM) and boosting \cite{schapire1998boosting}.
In addition to \cite{schapire1998boosting}, some more discussion exist for upper bound and margin of boosting \cite{wang2008margin,gao2013doubt} to which we refer the interested readers.

Assume we have training instances $\{(\b{x}_i, y_i)\}_{i=1}^n$ where $y_i \in \{-1,+1\}$ for binary classification.
The two classes may not be linearly separable. In order to handle this case, we map the data to higher dimensional feature space using kernels \cite{scholkopf2001learning,hofmann2008kernel}, hoping that they become linearly separable in the feature space. 
Assume $\b{h}(\b{x})$ is a vector which non-linearly maps data to the feature space.
Considering $\b{\alpha}$ as the vector of dual variables, the dual optimization problem \cite{boyd2004convex} in SVM is \cite{burges1998tutorial,schapire1998boosting}:
\begin{align}\label{equation_SVM_dual_optimization}
\underset{\b{\alpha}}{\text{maximize}} ~~~~ \underset{\{(\b{x}_i, y_i)\}_{i=1}^n}{\text{minimize}} ~~~~ \frac{y_i\, (\b{\alpha}^\top \b{h}(\b{x}_i))}{||\b{\alpha}||_2}.
\end{align}
Note that $y_i = \pm 1$ and $\b{\alpha}^\top \b{h}(\b{x}_i) \gtrless 0$; therefore, the sign of $y_i\, (\b{\alpha}^\top \b{h}(\b{x}_i))$ determines the class of the $i$-th instance.

On the other hand, the Eq. (\ref{equation_boosting_f_hat_normalized}) can be written in a vector form:
\begin{align}\label{equation_boosting_f_hat_normalized_vectorForm}
\widehat{f}(\b{x}_i) = \frac{\sum_{j=1}^k \alpha_j\, h_j(\b{x}_i)}{\sum_{j=1}^k \alpha_j} = \frac{\b{\alpha}^\top \b{h}(\b{x}_i)}{||\b{\alpha}||_1},
\end{align}
where $\b{h}(\b{x}_i) = [h_1(\b{x}_i), \dots, h_k(\b{x}_i)]^\top$ and $\b{\alpha} = [\alpha_1, \dots, \alpha_k]^\top$. Note that here, $h(\b{x}_i) = \pm 1$ and $\alpha_j$ is obtained from Eq. (\ref{equation_boosting_alpha_scaled}) or the line \ref{algorithm_AdaBoost_alpha} in Algorithm \ref{algorithm_AdaBoost}.

The similarity between the Eq. (\ref{equation_boosting_f_hat_normalized_vectorForm}) and the cost function in Eq. (\ref{equation_SVM_dual_optimization}) shows that boosting can be seen as maximizing the margin of classification resulting in a good generalization error \cite{schapire1998boosting}.
In other words, finding a linear combination in the high dimensional feature space having a large margin between the training instances of the classes is performed in the two methods.
Note that a slight difference is the type of norm which is interpretable because the mapping to feature space in boosting is only to $h(\b{x}_i) = \pm 1$ while in SVM, it can be any number where the sign is important. Therefore, $\ell_1$ and $\ell_2$ norms are suitable in boosting and SVM, respectively \cite{schapire1998boosting}.

Another connection between SVM (maximum margin classifier) and boosting is that some of the training instances are found to be most important instances, called \textit{support vectors} \cite{burges1998tutorial}. In boosting, also, weighting the training instances can be seen as selecting some \textit{informative} models \cite{freund1995boosting} which can be analogous to support vectors.

\subsection{Examples in Machine Learning: Boosting Trees and SVMs}

Both bagging and boosting are used a lot with trees \cite{quinlan1996bagging,friedman2001elements}.
The reason why boosting is very effective with trees is that trees have a large variance of estimation where the boosting and bagging can significantly reduce the variance of estimation as discussed before. 
Note that boosting is also used with SVM for imbalanced data \cite{wang2010boosting}.

\section{Conclusion}

This paper was a tutorial paper introducing overfitting, cross validation, generalized cross validation, regularization, bagging, and boosting. The theory behind these methods were explained and some examples of them in machine learning and computer vision were provided. 

\section*{Acknowledgment}
The authors hugely thank Prof. Ali Ghodsi (see his great online related courses \cite{web_classification,web_deep_learning}), Prof. Mu Zhu, Prof. Hoda Mohammadzade, Prof. Wayne Oldford, etc, whose courses have partly covered the materials mentioned in this tutorial paper.  

Sections \ref{section_mse}, \ref{section_overfitting}, \ref{section_CV_theory}, and some parts of Section \ref{section_regularization} (i.e., analysis of overfitting and regularization using SURE), and Section \ref{section_bagging_theory} (analysis of bagging), were primarily proposed by Prof. Ali Ghodsi verbally in his lectures, at University of Waterloo, available on YouTube\footnote{See \url{https://www.youtube.com/watch?v=21jL0I6wbns}.}. The credit of those sections is his. 
Moreover some parts of Section \ref{section_regularization} have been discussed in the books \cite{friedman2001elements} and \cite{goodfellow2016deep}.


\appendix

\section{Proof of Stein's Lemma}\label{section_stein_proof}

As the components of $\b{z} = [z_1, \dots, z_d]^\top \in \mathbb{R}^d$ are independent random variables with normal distribution, i.e., $z_i \sim \mathcal{N}(\mu_i, \sigma)$, we have:
\begin{align*}
f(\b{z}) &= f(z_1, \dots, z_d) \overset{(a)}{=} f(z_1) \times \dots \times f(z_d) \\
&= \prod_{i=1}^d \frac{1}{\sqrt{2 \pi \sigma^2}} \exp(-\frac{(z_i - \mu_i)^2}{2 \sigma^2}) \\
&= \frac{1}{\sqrt{(2 \pi \sigma^2)^d}} \exp(-\frac{\sum_{i=1}^d (z_i - \mu_i)^2}{2 \sigma^2}) \\
&= \frac{1}{\sqrt{(2 \pi \sigma^2)^d}} \exp(-\frac{||\b{z} - \b{\mu}||_2^2}{2 \sigma^2}),
\end{align*}
where $(a)$ is because $z_1 \indep \dots \indep z_d$.

We also have:
\begin{align*}
(\b{z} - \b{\mu})^\top\, \b{g}(\b{z}) = \sum_{i=1}^d (z_i - \mu_i)\, g_i.
\end{align*}

According to the definition of expectation, we have:
\begin{align*}
&\mathbb{E}\big((\b{z} - \b{\mu})^\top\, \b{g}(\b{z})\big) = \int_{\mathbb{R}^d} f(\b{z}) (\b{z} - \b{\mu})^\top\, \b{g}(\b{z})\, d\b{z} \\
&= \int_{\mathbb{R}^d} \frac{1}{\sqrt{(2 \pi \sigma^2)^d}} \exp(-\frac{||\b{z} - \b{\mu}||_2^2}{2 \sigma^2}) \sum_{i=1}^d (z_i - \mu_i)\, g_i\, \\
&\quad\quad\quad\quad\quad\quad\quad\quad\quad\quad\quad\quad\quad\quad\quad\quad\quad\quad dz_1\, \dots dz_d 
\end{align*}
\begin{align*}
&\overset{(a)}{=} \sigma^2 \sum_{i=1}^d \int_{\mathbb{R}^d} \frac{1}{\sqrt{(2 \pi \sigma^2)^d}} \exp(-\frac{||\b{z} - \b{\mu}||_2^2}{2 \sigma^2}) \frac{\partial g_i}{\partial z_i} \\
&\quad\quad\quad\quad\quad\quad\quad\quad\quad\quad\quad\quad\quad\quad\quad\quad\quad\quad dz_1\, \dots dz_d \\
&\overset{(b)}{=} \sigma^2 \sum_{i=1}^d \mathbb{E}(\frac{\partial g_i}{\partial z_i}),
\end{align*}
where $(a)$ uses integration by parts and $(b)$ is according to the definition of expectation. 
Q.E.D.

\bibliography{References}
\bibliographystyle{icml2016}

\end{document}